\documentclass[journal]{IEEEtran}
\usepackage{cite}

\usepackage[breaklinks]{hyperref}
\usepackage{xspace} 
\usepackage{mathrsfs}
\usepackage{amsmath,amsthm,amssymb,amsfonts}
\usepackage{booktabs}
\usepackage{makecell}
\usepackage{algorithmic}
\usepackage{array}
\usepackage{bm}
\usepackage{algorithm}
\usepackage[caption=false,font=normalsize,labelfont=sf,textfont=sf]{subfig}
\usepackage{textcomp}
\usepackage{stfloats}
\usepackage{url}
\usepackage[table]{xcolor}
\usepackage{xcolor}
\usepackage{verbatim}
\usepackage{graphicx}
\usepackage{cite}
\usepackage{array}
\usepackage{tabularx}
\usepackage{algorithm}
\usepackage{multirow}
\usepackage{bbding}
\usepackage{pifont}
\usepackage{wasysym}
\usepackage{color}
\usepackage{caption}
\usepackage{xcolor}
\usepackage{subcaption}
\usepackage{multicol}
\hypersetup{hidelinks,
	colorlinks=true,
	allcolors=black,
	pdfstartview=Fit,
	breaklinks=true}
\begin{document}
%
\title{Minds on the Move: Decoding Trajectory Prediction in Autonomous Driving with Cognitive Insights}

\author{Haicheng~Liao$^{*}$,
        Chengyue~Wang$^{*}$,
        Kaiqun Zhu,
        Yilong~Ren,
        Bolin~Gao,
        Shengbo~Eben Li, \IEEEmembership{Senior Member, ~IEEE},
        Chengzhong~Xu, ~\IEEEmembership{Fellow, ~IEEE}
        and Zhenning Li$^{\dag}$
\thanks{\dag\,Corresponding author; *\,Authors contributed equally.} \thanks{Haicheng Liao, Chengyue Wang, Kaiqun Zhu, Chengzhong Xu, and Zhenning Li are with the State Key Laboratory of Internet of Things for Smart City, University of Macau, Macau. Yilong Ren is with the School of Transportation Science and Engineering, Beihang University, Beijing, China. Bolin Gao and Shengbo Eben Li are with the School of Vehicle and Mobility, Tsinghua University, Beijing, China. E-mails: zhenningli@um.edu.mo. This research is supported by the  State Key Lab of Intelligent Transportation System under Project (2024-B001), Science and Technology Development Fund of Macau SAR (File no. 0021/2022/ITP, 0081/2022/A2, 001/2024/SKL), Shenzhen-Hong Kong-Macau Science and Technology Program Category C (SGDX20230821095159012), and University of Macau (SRG2023-00037-IOTSC).}}


%
%

\markboth{Journal of IEEE Transactions on Intelligent Transportation Systems, 2025}%
{Liao \MakeLowercase{\textit{et al.}}: Bare Demo of IEEEtran.cls for IEEE Journals}
%



\maketitle

\begin{abstract}
In mixed autonomous driving environments, accurately predicting the future trajectories of surrounding vehicles is crucial for the safe operation of autonomous vehicles (AVs). In driving scenarios, a vehicle's trajectory is determined by the decision-making process of human drivers. However, existing models primarily focus on the inherent statistical patterns in the data, often neglecting the critical aspect of understanding the decision-making processes of human drivers. This oversight results in models that fail to capture the true intentions of human drivers, leading to suboptimal performance in long-term trajectory prediction.
To address this limitation, we introduce a Cognitive-Informed Transformer (CITF) that incorporates a cognitive concept, Perceived Safety, to interpret drivers' decision-making mechanisms. Perceived Safety encapsulates the varying risk tolerances across drivers with different driving behaviors. Specifically, we develop a Perceived Safety-aware Module that includes a Quantitative Safety Assessment for measuring the subject risk levels within scenarios, and Driver Behavior Profiling for characterizing driver behaviors. 
Furthermore, we present a novel module, Leanformer, designed to capture social interactions among vehicles. CITF demonstrates significant performance improvements on three well-established datasets. In terms of long-term prediction, it surpasses existing benchmarks by 12.0\% on the NGSIM, 28.2\% on the HighD, and 20.8\% on the MoCAD dataset. Additionally, its robustness in scenarios with limited or missing data is evident, surpassing most state-of-the-art (SOTA) baselines, and paving the way for real-world applications.
\end{abstract}


%
\IEEEpeerreviewmaketitle

\section{Introduction}
%
%
%
%

\IEEEPARstart{I}{n} the evolving landscape of autonomous driving (AD) systems, the complex interactions between autonomous vehicles (AVs) and human-driven vehicles (HVs) present a significant challenge to achieving accurate trajectory prediction \cite{schwarting2019social,LIAO2024100116}.
The future trajectory of human-driven vehicles is essentially the result of the human driver's decision-making process \cite{chen2022intention,liao2024real}.
Since human drivers require reaction time to adjust their behavior when facing changes in the external environment \cite{arbabzadeh2019hybrid,li2024steering,liao2024and}, the dynamics of the vehicle will not change drastically in the short term, making \textbf{short-term} ($\leq2$ seconds) predictions relatively straightforward.
Nevertheless, long-term prediction necessitates models that accurately estimate the impact of numerous factors on the decision-making process of human drivers, a feat that is particularly challenging to achieve \cite{liao2024bat}. Recent advancements in algorithms and the availability of driving datasets have led to significant breakthroughs in trajectory prediction \cite{casas2020implicit}. However, the accuracy of long-term predictions (i.e., $>$2 seconds) remains a persistent challenge, primarily due to the inherent complexity of real-world driving scenarios. These challenges stem from the complex interactions between traffic agents, the impact of environmental factors like weather and road conditions, and the unpredictable nature of human driver behavior. These factors introduce significant uncertainty, making reliable long-term forecasts a persistent struggle for researchers in the field.

This backdrop prompts us to ask critical questions about the future trajectory of AD: Is the key to advancing AD not just in accumulating more data or refining algorithms, but in gaining a deeper understanding of the driving environment itself? How can we reshape our models to interpret and respond to the intricate human dynamics that underpin driving? Motivated by these questions, our research embarks on an innovative path. We propose a paradigm shift, extending beyond conventional data-driven approaches to embrace a critical yet often-neglected aspect of driving – the concept of \textbf{perceived safety}.

This concept, pivotal in shaping driving behaviors and decisions, is deeply rooted in psychological constructs, as detailed in \cite{rubagotti2022perceived}. According to the Theory of Planned Behavior, individual actions in driving are influenced by attitudes (driving behaviors towards others), subjective norms (personal evaluation of safety), and perceived behavioral control (confidence in driving ability) \cite{ajzen1991theory}. Further depth is added by neuroscientific research, such as studies by \cite{10468619,liao2024less} and \cite{kronemer2022human}, which unveil that perceived safety is an intricate blend of both conscious and instinctive responses, involving the amygdala's emotional processing and the prefrontal cortex's rational decision-making.
Notably, this nuanced understanding of perceived safety is exemplified in diverse driving scenarios. For instance, when encountering a close car ahead, different drivers exhibit markedly varied responses. An aggressive driver, possibly influenced by sensation-seeking tendencies \cite{zuckerman1990psychophysiology}, might quickly swerve, perceiving lower risk. Conversely, a cautious driver, perhaps more risk-averse \cite{rabin2013risk}, might opt for a complete stop. These behaviors, far from being random, are intricately linked to each driver's psychological profile and past experiences, revealing a significant limitation in current AD systems: their inability to account for these complex, cognitive behavioral patterns. Overall, perceived safety and its influence on the decisions of drivers with different behaviors.

{
In response, our research introduces the Cognitive-Informed Transformer (CITF) that integrates the concept of perceived safety into trajectory prediction for AVs. This integration does more than add a new variable; it injects a human-centric perspective into the heart of these systems. By doing so, we aim to enhance the models' ability to interpret driving behaviors, leading to optimal long-term predictions. This approach, promises a transformative impact on the predictive capabilities of AD systems, aligning them more closely with the multifaceted nature of human driving behavior. 

Overall, the key contributions of this study include:
\begin{itemize}
    \item We introduce the Quantitative Safety Assessment (QSA) as a cornerstone component for objectively evaluating the safety of driving scenarios. In addition, we establish Driver Behavior Profiling (DBP) upon the QSA framework to differentiate between distinct driver profiles. This DBP effectively captures and interprets continuous nuances in driving behavior, while eliminating the dependence on manual labeling or predefined time windows.

\item We introduce an innovative module, named Leanformer, that represents a significant advancement in understanding social interactions on the road. This lightweight transformer-based framework is adept at capturing the subtle and complex inter-vehicular interactions that occur in everyday traffic. This development reflects a paradigm shift in AD research, aligning with the latest advancements and understanding of vehicular social dynamics.

\item CITF significantly outperforms the SOTA baseline models when tested on the NGSIM, MoCAD, and HighD datasets. It maintains impressive performance even when trained on only 25\% of the dataset and with a much smaller number of model parameters, demonstrating its efficiency and adaptability in various traffic scenes, including highways, campuses, and busy urban locales. Importantly, in a significant stride towards practical applicability, CITF shows unparalleled resilience in scenarios with incomplete or inconsistent data. 
\end{itemize}

}

\section{Related Work}\label{Related work}
\textbf{Trajectory Prediction For Autonomous Driving.} {
In the field of trajectory prediction, the analysis of prediction performance is often categorized into short-term and long-term horizons \cite{chen2022intention, wang2025wake}. Early research employed physical models to represent vehicle motion dynamics, thereby estimating future trajectories. In \cite{brannstrom2010model}, a trajectory prediction model based on the bicycle model was proposed and successfully applied to an accident warning system. While physical models achieved significant progress in short-term prediction horizons, their inherent simplicity limited their performance in long-term predictions \cite{wang2024dynamics}. The complexity of human driving behavior, influenced by numerous factors such as cognitive processes, interactions with surrounding vehicles, and environmental conditions, renders long-term prediction a particularly challenging task \cite{wang2025wake,liao2024human2}. In response to these challenges, researchers have begun integrating deep learning models to account for these factors in the trajectory prediction process.
Notable prior efforts \cite{wang2023wsip, ijcai2024p656,liao2024bat} have explored the complex social dynamics among traffic participants, revealing crucial latent insights that enhance predictive accuracy. Transformer-based models \cite{ijcai2024p756,wang2024nest} have been increasingly employed for their ability to predict future trajectory distributions effectively. Graph Neural Networks (GNNs) are also gaining traction for capturing dynamic interactions in complex traffic scenes \cite{rowe2023fjmp,ijcai2024p657}. These approaches primarily focus on understanding the temporal and spatial interplays between traffic agents from historical data to optimize accuracy. Generative models \cite{ijcai2024p811}, including Variational Auto Encoders (VAEs), Diffusion models, and Generative Adversarial Networks (GANs), are also being explored for their potential to generate multiple future trajectory possibilities from latent distributions, offering a probabilistic perspective of future paths in this field.}

\textbf{Perceived Safety Concept.} The notion of perceived safety has been a focal point in psychology and physical human-robot interaction (pHRI) studies \cite{guiochet2017safety}. In pHRI, it is crucial for assessing and representing individuals' perceptions of danger and comfort during interactions with autonomous systems like mobile robots \cite{chen2020yard}, industrial manipulators \cite{davis2023role}, humanoid robots \cite{busch2019evaluation} and AVs \cite{sun2021joint}. Despite its relevance, perceived safety remains a challenging concept to quantify due to its subjective nature \cite{bartneck2009measurement}. Our study breaks new ground in this area by proposing a novel quantitative criterion for perceived safety in self-driving trajectory prediction, drawing from Safety State Metrics (SSMs) and human decision-making processes. This innovation enables our model to more accurately interpret driving behavior and traffic conditions, thereby enhancing prediction accuracy in mixed autonomy environments.

\textbf{Driving Behavior Understanding.} Existing studies in driving behavior have formulated various criteria and metrics for detecting and representing driving patterns, using scales like the Social Value Orientation (SVO) \cite{murphy2011measuring}, Driving Anger Scale (DAS) \cite{deffenbacher1994development}, among others \cite{taubman2004multidimensional}. While these methods have been successful, as noted by \cite{schwarting2019social} and \cite{chandra2020cmetric}, they typically depend on manually annotated labels and predetermined sliding time windows for analysis. Our research diverges from these traditional approaches by proposing a dynamic, adaptive set of behavior-aware criteria. This model captures driving behavior in real-time through continuous behavioral data representation, eliminating the reliance on manual labeling in the training phase. This novel approach not only offers enhanced flexibility over fixed-category methods but also effectively addresses the challenges of label shifts and time window selection, leading to a more accurate and fluid representation of driving behavior. This advancement significantly contributes to the development of more refined and effective behavior prediction methodologies in autonomous driving systems.
\begin{table*}[htbp]
\centering
\caption{Primary notations and their meanings.}
\begin{multicols}{2}
\begin{tabularx}{\columnwidth}{cX}
    \toprule
\textbf{Notation} & \textbf{Meaning} \\
 \midrule
$\bm{X}_{0}^{t-t_{h}:t}$ & Historical states of the target vehicle within a defined duration $t_{h}$\\
$\bm{X}_{1:n}^{t-t_{h}:t}$ & Historical states of surrounding agents $1$ to $n$ within the duration $t_{h}$ \\
$\bm{Y}_{0}^{t: t+t_{f}}$  & Predicted future trajectory of target vehicle over the ensuing \( t_{f} \) time intervals\\
$p_{0:n}^{t-t_{h}:t}$ & x- and y- coordinates of the target vehicle and surrounding agents from time horizon \( t-t_{h} \) to \( t \)\\
$v_{0:n}^{t-t_{h}:t}$ & Velocity of the target vehicle and its surrounding agents from \( t-t_{h} \) to \( t \) \\
$t_h$ & Acceleration of the target vehicle and its surrounding agents from \( t-t_{h} \) to \( t \) \\
$n$ & Number of surrounding vehicles \\
$M$ & Total number of the predicted potential trajectories\\
$\hat{\bm{x}}_i$ & Original longitudinal coordinates of vehicle $i$ \\
$\bm{M}$ & Probability of different maneuvers\\
$\tau_{\mathrm{sc}}$ & Given time threshold $i$ \\
$a_{i}$ & Acceleration of the traffic agent $i$ \\
$v_{x}^{t}$ & Lateral velocity \\
$v_{y}^{t}$ & Longitudinal position velocity at time $t$\\
$p_{x}^{t}$ & Longitudinal position coordinates at time $t$\\
$p_{y}^{t}$ & Horizontal coordinate\\
$\mathcal{S}_{i}^{t}$ &Safe Magnitude Index for the $i$-th traffic agent at time $t$ \\
$\bm{H}$ & Set of safety indices\\
$I_N$ & Identity Matrix\\
$A$ & Adjacency matrix\\
$\tilde{\mathbf{A}}$ & Degree matrix for normalizing the graph structure \\
$\mathbf{Z}^{k+1}_i$   & Learned feature matrix the $i$-th agent from GCNs \\
 ${\alpha}^{\textit{behavior }}$ & Output generated by the multi-head self-attention mechanism for DBP\\
  ${\alpha}^{\textit{priority }}$ & Output of the multi-head self-attention mechanism for the Priority-Aware Module\\
\bottomrule
\end{tabularx}

\begin{tabularx}{\columnwidth}{cX}
\toprule
\textbf{Notation} & \textbf{Meaning} \\
 \midrule
 $h_s$ & Total number of attention heads for QSA\\
 ${\alpha}^{\textit{safety }}$ & Output of the multi-head self-attention mechanism for QSA\\
${G}^{t}$ & Dynamic geometric graph at time step $t$\\
$V^{t}$ & Node set of the DGG at time step $t$\\
${v}_{i}^{t}$ & $i$-th node of the DGG at time step $t$\\
$v_{i}^{t}$ & Edge of the $i$-th node at time step $t$\\
$d(v_{i}^{t}, v_{j}^{t})$ & Shortest distance between $i$-th and $j$th node\\
$\mathcal{N}_{i}^{t}$ &Neighborhood set of $i$-th node at time step $t$\\
${\mathcal{J}}_{i}^{{t}}$ &Behavior-awre criteria of $i$-th node at time step $t$\\
${J}_{i}^{t}(D)$ & Degree centrality of $i$-th node at time step $t$\\
${J}_{i}^{t}(C)$ & Closeness centrality of $i$-th node at time step $t$\\
${J}_{i}^{t}(E)$ & Eigenvector centrality of $i$-th node at time step $t$\\
${J}_{i}^{t}(B)$ &Betweenness centrality of the $i$-th node at time step $t$\\
${J}_{i}^{t}(P)$ &Power centrality of $i$-th node at time step $t$\\
${J}_{i}^{t}(K)$ & Katz centrality of $i$-th node at time step $t$\\
$\left| \mathcal{N}_{i}^{t}\right|$ & Total elements in Neighborhood set $\mathcal{N}_{i}^{t}$ \\
$\sigma_{j,k}$ & Total number of shortest paths between $v_{j}^{t}$ and $v_{k}^{t}$ at time step $t$\\
$\sigma_{j,k}(v_{i})$ & Number of the paths traversing $v_{i}^{t}$ at time step $t$\\
$A^{k}_{ii}$ & $i$-th diagonal element of the adjacency matrix to the $k$-th power\\
 $k!$ & Factorial of $k$\\
$\alpha^{k}$ & Decay factor\\
$\beta^{k}$ & Weight for immediate neighbor nodes\\
$d_{k}$&Dimensionality of the projected key vectors for the Leanformer framework\\
${\bar{O}_{\textit{safety}}}^{t-t{h}:t}$ & Safety feature output from the QSA\\
${\bar{O}_{\textit{behavior}}}$ & Behavior feature output from the DBP\\
${\bar{O}}$ & Output from the Interaction-Aware Module\\
${{O}}_{\textit{priority}}^{t-t_{h}:t}$ & Behavior feature output from the Priority-Aware Module\\
$\mathbf{Y}^{\text{pred}}(T)$ & Predicted trajectories at the prediction horizon $T$\\
$\mathbf{Y}^{\text{gt}}(T)$ & Ground-truth trajectory at the  horizon $T$\\
 \bottomrule
\end{tabularx}
\end{multicols}
\label{table_variables}%
\end{table*}

\begin{figure*}[t]
  \centering
\includegraphics[width=0.93\textwidth]{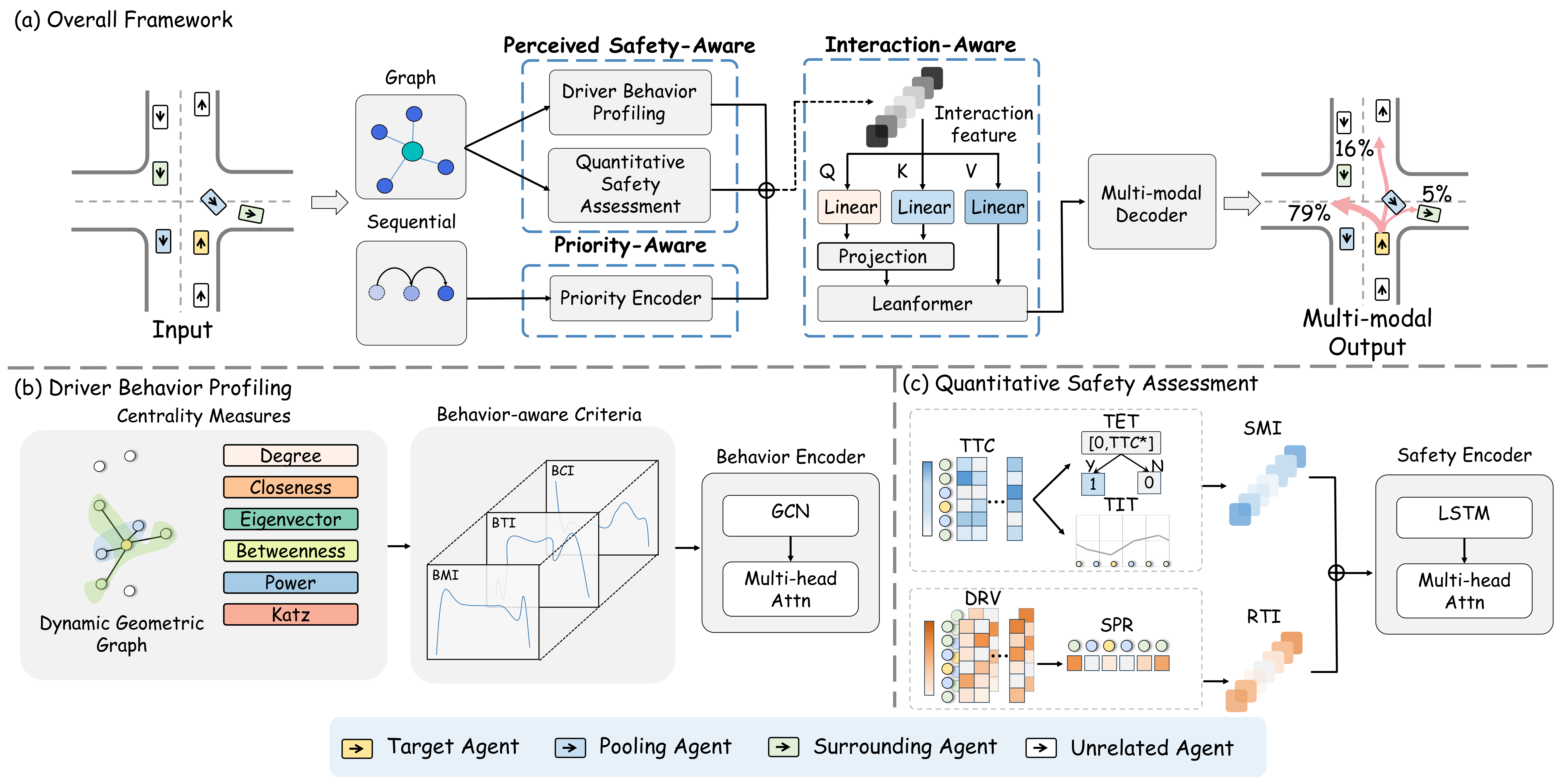} 
  \caption{Pipeline of CITF. It is an encoder-decoder model (a) and includes four essential parts: Perceived Safety-Aware Module generates both safety and behavior features through driver behavior profiling (b) and quantitative safety assessment (c) components, respectively. These features, along with the priority feature derived from the Priority-Aware Module, are integrated into the Interaction-Aware Module, embedded by the Leanformer framework. Finally, this integration results in a high-level fusion, which is then fed into the Multimodal Decoder to produce a multimodal prediction distribution for the target vehicle.}
  \label{fig1} 
\end{figure*}

\section{Problem Formulation}\label{Problem Formulation}
In mixed autonomy traffic scenarios, trajectory prediction models within AVs are tasked with forecasting the future trajectories of all surrounding vehicles within their perception range. According to surveys by Mozaffari et al. \cite{mozaffari2020deep} and Ding et al. \cite{ding2023incorporating}, the single-agent prediction setting remains a prevalent approach in the field of trajectory prediction. In this setting, the model is developed by selecting one vehicle from the surrounding vehicles as the prediction target. During the evaluation phase, the model's predictive capability is assessed in a traversal manner, which treats each vehicle in the scene as the prediction target once. Adhering to this setting, we can define the terminology used in our study as follows: 
\begin{itemize}
\item {Target vehicle:}  The vehicle is designated as the subject of the trajectory prediction task.

\item {Surrounding agents:} The AV and all of its perceived traffic agents, excluding the target vehicle.

\end{itemize}
In summary, our problem could be formulated as developing a trajectory prediction model that could utilize the historical states (position, velocity, etc.) of both the target vehicle $\bm{X}_{0}^{t-t_{h}:t}$ and the surrounding vehicles $\bm{X}_{1:n}^{t-t_{h}:t}$ spanning from time $t-t_{h}$ to present moment $t$, to predict the future trajectory $\bm{Y}_{0}^{t: t+t_{f}}$ of the target vehicle over the ensuing $t_f$ time intervals.

\subsection{Discretized Inputs and Outputs}
Theoretically, the inputs (historical states) and outputs (future trajectories) should be represented in a continuous form. However, in practical deployment, the sensors on AVs collect data at fixed intervals. Therefore, to maintain consistency with the collected data, it is widely accepted in both academia \cite{chen2023stochastic, zhou2023csir} and industry \cite{chang2019argoverse} to use discretized inputs and outputs when developing trajectory prediction models. Specifically, we define the inputs and outputs as follows:
\begin{itemize}
\item {Inputs:} The historical states $\bm{X}_{0:n}^{t-t_{h}:t}$ of the target vehicle and its surrounding agents, consists of a sequence of historical states $\{ \bm{X}_{0:n}^{t-t_{h}}, \bm{X}_{0:n}^{t-t_{h}+1}, ..., \bm{X}_{0:n}^{t} \}$. At any time $t$, the historical states $\bm{X}_{0:n}^t$ comprise 2D position coordinates \( p_{0:n}^{t} \), velocity \( v_{0:n}^{t} \), and acceleration \( a_{0:n}^{t} \).

\item {Outputs:} The predicted trajectory of the target vehicle, denoted as $\bm{Y}_{0}^{t:t+t_{f}}$, consists of a sequence of predicted positions $\{\bm{p}_{0}^{t+1},\bm{p}_{0}^{t+2},\ldots,\bm{p}_{0}^{t+t_{f}-1},\bm{p}_{0}^{t+t_{f}}\}$. 
\end{itemize}

For brevity, we also list the primary notations and their meanings in Table \ref{table_variables}. 

\subsection{Multi-modal Probabilistic Maneuver Prediction}
 we adopt a multimodal prediction framework to tackle the inherent uncertainty and variability in predictions. By evaluating different possible maneuvers that the target vehicle might perform, the framework computes the probability of each maneuver based on historical states $\bm{X}_{0:n}^{t-t_{h}:t}$, which include 2D position coordinates, velocity, and acceleration over a defined time horizon \(t_h\).  This approach generates multiple predictions while also quantifying the confidence level associated with each prediction. This allows AVs to account for and respond to the uncertainty inherent in prediction outcomes, providing a valuable advantage for decision-making processes.

In this study, we employ a hierarchical Bayesian framework to predict future trajectories. At each time step, we evaluate the probability distribution over the possible maneuvers $\bm{M}$ of the target vehicle. To capture the driver's behavioral nuances, we decompose the vehicle’s potential maneuvers into two distinct sub-maneuvers: $\bm{M} = (\bm{M}_{p}, \bm{M}_{v})$. Here, $\bm{M}_{p}$ represents the position sub-maneuver, encompassing three discrete driver decisions: left lane change $m_{l}$, right lane change $m_{r}$, and lane keeping $m_{k}$. Similarly, $\bm{M}_{v}$ denotes the speed sub-maneuver, with three options: accelerating $m_{a}$, braking $m_{b}$, and maintaining constant speed $m_{c}$.

Following this categorization, the framework generates detailed trajectories for the vehicle conditioned on each maneuver within a predefined distributional form.
At each current time $t$, we extend the trajectory prediction task to compute the plausible trajectory distribution $\bm{P}\left(\bm{Y}_{0}^{t:t+t_{f}} \mid \bm{M}, \bm{X}_{0:n}^{t-t_{h}:t}\right)$. In particular, given the estimated maneuvers $\bm{M}$, the probability distribution of the multimodal trajectory predictions $\bm{Y}_{0}^{t:t+t_{f}}$ is parameterized as a bivariate Gaussian distribution with the estimable parameters  $\bm{\Omega}$:

\vspace{-9pt}
{\begin{small}
\begin{flalign}\label{eq.5}
     \bm{P}\left(\bm{Y}_{0}^{t:t+t_{f}} \mid \bm{M}, \bm{X}_{0:n}^{t-t_{h}:t}\right)
     &= \bm{P}_{\bm{\Omega}} (\bm{Y}_{0}^{t:t+t_{f}} \mid \bm{M}, \bm{X}_{0:n}^{t-t_{h}:t}) \\\nonumber
     &= \bm{N}(\bm{Y}_{0}^{t:t+t_{f}}|\mu(\bm{X}_{0:n}^{t-t_{h}:t}),\Sigma(\bm{X}_{0:n}^{t-t_{h}:t}))
\end{flalign}
\end{small}}

Here, $\bm{\Omega}=\left[\Omega^{t+1}, \ldots,\Omega^{t+t_{f}}\right]$, and $\Omega^{t}=[\mu^{t},\Sigma^{t}]$ represents the mean and variance of the distribution of predicted trajectory point at time $t$. Correspondingly, the multi-modal predictions are then formulated as a Gaussian Mixture Model:

\vspace{-9pt}
{\begin{footnotesize}
\begin{flalign}\label{eq.6}
&\bm{P}\left(\bm{Y}_{0}^{t:t+t_{f}} \mid \bm{M}, \bm{X}_{0:n}^{t-t_{h}:t}\right) \\\nonumber&=\sum_{\forall i} \bm{P}\left(\bm{M}_{i} \mid \bm{X}_{0:n}^{t-t_{h}:t}\right) \bm{P}_{\bm{\Omega}}\left(\bm{Y}_{0}^{t:t+t_{f}} \mid \bm{M}_{i},\bm{X}_{0:n}^{t-t_{h}:t}\right)
\end{flalign}
\end{footnotesize}}
where $M_{i}$ denote the $i$-th element in possible maneuvers $\bm{M}$.

\section{Trajectory Prediction Model}\label{Proposed Model}
Figure \ref{fig1} shows the hierarchical framework of CITF. Rooted in the encoder-decoder paradigm, the model seamlessly incorporates four novel modules: the Perceived Safety-Aware Module, the Priority-Aware Module, the Interaction-Aware Module, and the Multimodal Decoder. Collectively, these modules are designed to capture human-machine interactions between the target vehicle and its surrounding agents and emulate the human decision-making process during driving.  Detailed overviews of these modules follow.

\subsection{Perceived Safety-Aware Module}
As mentioned before, perceived safety \cite{nair2021sharing} plays a critical role in human decision-making during driving. The nuances in perceived safety can significantly affect human driver behavior and further impact AV's inability to account for the complex, cognitive behavioral patterns in mixed autonomy environments.
Recognizing this, the Perceived Safety-Aware Module is introduced to establish precise criteria and quantify specific criteria for evaluating perceived safety. It consists of two integral components: \textbf{1) Quantitative Safety Assessment:} This component focuses on the development of physically based, measurable criteria that can accurately reflect how humans subjectively assess the level of danger; \textbf{2) Driver Behavior Profiling:} This component aims to provide in-depth, real-time analysis and profiling of the continuous driving behavior of human drivers especially those influenced by their perceived safety.
Together, as shown in Table \ref{table_0}, these components are meticulously designed to enhance AVs' understanding of perceived safety in driving contexts, allowing them to better understand and anticipate human driver responses.
By incorporating this type of valuable prior knowledge, we facilitate the synthesis of human-like contextual patterns for the proposed model. This enhancement, along with the Priority-Aware Module, allows our Interaction-Aware Module to better decipher and assimilate the intentions of traffic agents and more closely match the intricacies of human cognition and decision-making in driving scenarios, resulting in improved overall model performance.

\subsubsection{Quantitative Safety Assessment} 
As shown in Figure \ref{fig1} (c), this component $\bm{H}$ includes two safety indices: the Safe Magnitude Index (SMI) and the Risk Tendency Index (RTI).  In a nutshell, the SMI focuses primarily on quantifying the spatio-temporal distance between different agents and the possibility of collision to evaluate the absolute safety level in real-time scenarios. Conversely, RTI tends toward a more subjective analysis, capturing dynamic shifts in safety trends and congestion conditions that reveal potential escalation or mitigation of risk over time.

\begin{table*}[htbp]
  \centering
  \caption{Perceived-Safety criteria and their interpretations.}
  \resizebox{0.9\linewidth}{!}{
    \begin{tabular}{ccc}
    \toprule
    Indies & Sub-indicators & Definition \\
    \midrule
    \multirow{5}[0]{*}{Safe Magnitude Index (SMI)} &  \makecell{Time-to-Collision (TTC)} &  \makecell{Time until a potential collision occurs  between the agent \\and another agent moving at the current velocity} \\
    \cmidrule{2-3} 
          &  \makecell{Time Exposed Time-to-Collision (TET)} &  \makecell{Cumulative sum of instances in which a driver approaches\\ a leading vehicle with a TTC below a predefined threshold value} \\
           \cmidrule{2-3} 
          &  \makecell{Time Integrated Time-to-Collision (TIT)} &  \makecell{Integral of the TTC profile over time if it remains below a specified threshold} \\
        \midrule
    \multirow{2.5}[0]{*}{Risk Tendency Index (RTI)} &  \makecell{Subjective Risk Perception (SRP)} &  \makecell{Congestion level of each agent in complex traffic environment} \\
    \cmidrule{2-3} 
          &  \makecell{Dynamic Risk Volatility (DRV)} &  \makecell{Rate of change of congestion level for each agent in a complex traffic environment} \\
    \bottomrule
    \end{tabular}%
    }
  \label{table_0}%
\end{table*}%

\textbf{Safe Magnitude Index.} In the traffic safety domain, three metrics in SSMs have gained prominence for their comprehensive portrayal of on-road risks: Time-to-Collision, Time Exposed Time-to-Collision (TET), and Time Integrated Time-to-Collision (TIT) \cite{minderhoud2001extended}. Originating from traffic conflict studies, these metrics are essential tools in microscopic traffic simulations to assess traffic safety. Correspondingly, we synthesize TTC, TIT, and TET into a ternary composite structure within SMI. This composite structure is introduced to evaluate the dynamics of interaction between traffic agents and to estimate the likelihood of potential collisions for each vehicle in real-time scenarios. Specifically, the SMI for an agent at a specific time ${t}$ can be expressed as $
 {\mathcal{S}}_{i}^{{t}}=\left[{ TTC^{{t}}_{i}},{TET^{{t}}_{i}}, { TIT^{{t}}_{i}}\right]$.
To align these with the traffic scenarios, we made slight modifications. Using the 2D position coordinates \( p_{i}^{t} \), \( p_{j}^{t} \) and velocity \( v_{i}^{t} \), \( v_{j}^{t} \) for vehicles \( i \) and \( j \) at time \( t \).

\textbf{1) Time-to-Collision:}
TTC is a widely accepted measure used to evaluate the time available before two vehicles collide if they continue on their current trajectories. It offers insights into imminent collision risks and serves as an early warning indicator. The TTC for the \( i \)-th vehicle is computed as $TTC_{i}^{t} = -\frac{d_{i,j}^{t}}{\dot{d}_{i,j}^{t}}$, where \( d_{i,j} \) represents the distance between vehicles \( i \) and \( j \), and \( \dot{d}_{i,j} \) is its rate of change:
\begin{equation}\label{eq.4_1}
    \left\{\begin{array}{l}
d_{i,j}^{t} =\sqrt{\left({p}_i^{t}-{p}_j^{t}\right)^{\top}\left({p}_i^{t}-{p}_j^{t}\right)} \\
\dot{d}_{i,j}^{t} =\frac{1}{d_{i,j}^{t}}\left({p}_i^{t}-{p}_j^{t}\right)^{\top}\left({v}_i^{t}-{v}_j^{t}\right)
\end{array}\right.
\end{equation}
Accordingly, the higher the TTC value, the lower the risk of collision for the vehicle in this case.

\textbf{2) Time Exposed Time-to-Collision:}
TET measures the exposure duration to critical TTC values within \( t_h \). It is the sum product of a switching variable and a time threshold $\tau_{\mathrm{sc}}$ (set at 0.1s):
$TET_i^{t_{k}}=\sum_{{t_{k}}=t-t_{h}}^{t} \delta_i(t_{k}) \cdot \tau_{\mathrm{sc}}$
with the switching variable given by:
\begin{equation}
\delta_i(t_{k})=\left\{\begin{array}{lc}
1 & \forall \ \ \ 0  \leq \mathrm{TTC}_i^{t_{k}} \leq \mathrm{TTC}^*\\
0 & \text { otherwise }
\end{array}\right.
\end{equation}
In our study, \( \mathrm{TTC}^* = 3.0s \), delineating safety threshold.

\textbf{3) Time Integrated Time-to-Collision:}
An adaptation of TET, TIT integrates the TTC profile to evaluate safety levels. It factors in the evolution of each vehicle's TET temporally:
\begin{equation}
TIT_i^{{t}_{k}}=\sum_{{t_{k}}=t-t_{f}}^{t_{h}}\left[\mathrm{TTC}^*-\mathrm{TTC}_{i}({t}_{k})\right] \cdot \tau_{\mathrm{sc}}
\end{equation}
Elevated values of TTC, TET, and TIT imply sustained exposure to potential collision risks, underscoring a deterioration in perceived safety. Overall, the SMI provides both real-time crash risk assessment and an aggregated risk evaluation over a defined period, eliminating the need for historical crash data. It also takes into account the fluctuation and rate of change of these risks, assessing the safety benefits of AVs in mixed autonomy environments, and offering a comprehensive safety evaluation for each agent.

\textbf{Risk Tendency Index.} To further capture congestion patterns in complex traffic environments, we propose an index between the $i$-th and $j$-th vehicles at time $t$, denoted as subjective risk perception indicator (SPR), i.e. ${ {R}}_{i}^{{t}}$, and dynamic risk volatility indicator (DRV), i.e. $ {R}^{{t}}_{i,j}$, respectively:
\begin{equation}\label{eq.12}
{ {R}}_{i}^{{t}}= {R}_{j}^{{t}}=\left[ \log({R}^{{t}}_{i,j}),  \log({\dot{R}}^{{t}}_{i,j})\right]^{T}, \forall i, j \in[0,n], i \neq j 
\end{equation}
In this context, the vector ${R}^{{t}}_{i,j}$ with larger values indicates an increased risk of collision, while the vector $\dot{ {R}^{{t}}}_{i,j}$ characterizes the dynamic congestion conditions in complex traffic scenarios.
Then, the set of the safety indices $\bm{H}=\{\mathcal{S}_{0}^{t-t_{h}:t},\mathcal{R}_{0}^{t-t_{h}:t},\ldots,\mathcal{S}_{i}^{t-t_{h}:t},\mathcal{R}_{i}^{t-t_{h}:t}\ \forall i \in[1,n]\}$ serve as contextual cues and are then fed into the safety encoder for embedding into high-level safety features. 
The definitions of SPR and DRV are defined as follows:
\begin{equation}
 {R}^{{t}}_{i,j}= {R}^{{t}}_{j,i}=\left\{\begin{array}{l}
1 / e^{{q}_{i,j}^{{t}}}, {q}_{i,j}^{{t}}>0 \\
0, {q}_{i,j}^{{t}}=0
\end{array} \right. 
\end{equation}
where the DRV $ {\dot{R}}^{{t}}_{i,j}$ represents the gradient to evaluate fluctuations in SPR $ {{R}}^{{t}}_{i,j}$ and can be expressed as follows:
\begin{equation}
\dot{ {R}^{{t}}}_{i,j}=\dot{ {R}}^{{t}}_{j,i}=\left\{\begin{array}{l}
1 / e^{\dot{{q}}_{i,j}^{{t}}}, \dot{{q}}_{i,j}^{{t}}>0 \\
0, \dot{{q}}_{i,j}^{{t}}=0
\end{array}\right. 
\end{equation}
The quantities ${q}_{i,j}^{{t}}$ and $\dot{q}_{i,j}^{{t}}$ are calculated based on several critical parameters related to the dynamics of two traffic agents. These parameters include the lateral velocity $v_{x}^{t}$, longitudinal velocity $v_{y}^{t}$, 2D position coordinate $ p_{x}^{t}$ and $ p_{y}^{t}$,  as well as the lateral  $a_{x}^{t}$and longitudinal $a_{y}^{t}$. Mathematically, it can be represented as follows:
\begin{equation}
   {q}_{i,j}^{{t}} = \max \left(-\frac{\Delta_{i,j} v_{x}^{t} \times \Delta_{i,j} p_{x}^{t} +\Delta_{i,j} v_{y}^{t} \times \Delta_{i,j} p_{y}^{t}}{\Delta_{i,j} v_{x}^2+\Delta_{i,j} v_{y}^2}, 0\right)
\end{equation}
\begin{equation}
     \dot{q}_{i,j}^{{t}} = -\frac{\Delta_{i,j} a_{x}^{t} \times \Delta_{i,j} p_{x}^{t} +\Delta_{i,j} a_{y}^{t} \times \Delta_{i,j} p_{y}^{t}}{\Delta_{i,j} a_{x}^2+\Delta_{i,j} a_{y}^2}
\end{equation}
where the $\Delta_{i,j} (\cdot)$ denotes the difference between quantities of the $i$-th and $j$-th vehicles.
A larger vector $ {R}^{{t}}_{i,j}$ indicates a higher risk of collision, while the vector $\dot{ {R}^{{t}}}_{i,j}$ describes the dynamic congestion conditions in complex traffic scene.

\textbf{Safety Encoder.}
This encoder applies the GCNs \cite{hamilton2017inductive} to analyze the spatial layouts of traffic agents and their environmental context. Next, it enhances the scaled dot-product multi-head self-attention mechanism \cite{vaswani2017attention} for a nuanced analysis of temporal relationships within safety indices. 

Specifically, for GCN, we employ a convolutional neural network on a fully connected interaction multigraph to capture the dynamic geometric relationships among traffic agents. This multigraph operational layer sequentially incorporates the set of safety indices ${\bm{H}}$ as nodes. These nodes represent various security-related properties and states of the traffic agents over time.  To establish the connections between these nodes, we use an adjacency matrix $A$, which is detailed in the following subsection. This matrix represents the edges of the graph and is critical in defining the interactions and relationships between different nodes (agents) within the graph. Formally,
\begin{equation}
\bm{Z}^{k+1}_i=\phi_{\text {ReLU}}\left(\tilde{\bm{D}}^{-\frac{1}{2}} \tilde{\bm{A}} \tilde{\bm{D}}^{-\frac{1}{2}} \bm{Z}^k_i \bm{W}^k_i\right)
\end{equation}
where the matrix $\tilde{\mathbf{D}}$ serves as the scale factor of $\tilde{\mathbf{A}}$, is the degree matrix for normalizing the graph structure.
It helps to balance the influence of each node based on its connectivity. The $\bm{W}^k_i$ represents the trainable weight matrix of the GCN for the $k$-th layer, while $\phi_{\text {ReLU}}$ is the Rectified Linear Unit (ReLU) activation function. Consequently,  the matrix $\tilde{\mathbf{A}}$ can be defined as $\bm{\tilde{A}}=A+\lambda_{A} I_N$,
where $\lambda_{A}$ is the weight and $I_N$ is the identity matrix. The output of the $k$-th convolutional layer, denoted as $\mathbf{Z}^{k+1}_i$, represents the learned feature matrix of the $i$-th agent.
Moreover, the initial feature $\bm{Z}^{0}_i=\phi_{\text {MLP}} (\bm{H})$, where $\phi_{\text {MLP}}$ denotes a Multi-Layer Perceptron (MLP). The MLP serves as a fully connected layer to embed the safety indices $\bm{H}$ into a feature space suitable for graph convolution. In addition, we employ a tri-layer convolutional neural network that incorporates scatter and gathers operations to parallelize the learning of contextual information and spatio-temporal agent interdependencies.

Next, the feature matrix $\mathbf{Z}^{k+1}_i,\mathbf{Z}^{k}_i$ and $\mathbf{Z}^{k-1}_i$ output from the $(k+1), k$ and $(k-1)$-th GCNs is then converted to the query, key, value vectors, respectively, by the multi-head self-attention mechanism within the encoder to produce the high-level safety features. Formally,
\begin{equation}
\begin{cases}
{Q}_i^{\textit {safety }}={W}^{Q^{\textit {s }}}\phi_{\textit{MLP}}\left(\mathbf{Z}^{k+1}_i\right) \\
{K}_{i}^{\textit {safety }}={W}^{K^{\textit {s }}} \phi_{\textit {MLP}}\left(\mathbf{Z}^{k}_i\right)\\
{V}_{i}^{\textit {safety }}={W}^{V^{\textit {s }}} \phi_{\textit {MLP}}\left(\mathbf{Z}^{k-1}_i\right)
\end{cases}
\end{equation}
where the ${W}^{Q^{\textit {s }}}, {W}^{K^{\textit {s }}}, {W}^{V^{\textit {s }}}$ are learnable weights that can be optimized via gradient descent. For the $i$-th self-attention head $ {head}_i$, the formulation is as follows:
\begin{equation}
\textit{head}_i^{\textit{ s }}= { \phi_{\textit {softmax}}}\left(\frac{{Q}_i^{\textit {safety }} ({K}_i^{\textit {safety }})^{\top}}{\sqrt{d_s}}\right) {V}_i^{\textit {safety }}
\end{equation}
In the equation provided, $ { \phi_{\textit {softmax}}}(\cdot)$ denotes the softmax activation function, while ${d_s}$ represents the dimensionality of the projected key vectors. The output generated by the self-attention mechanism can be expressed as ${\alpha}^{\textit{safety }}  = \sum_{i=1}^{h_s} {\textit{head}_i^{\textit{ s }}}$, where $h_s$ is the total number of attention heads.

To increase training stability and efficiency, our model takes inspiration from ResNet \cite{he2016deep} and incorporates Gated Linear Units (GLUs) \cite{dauphin2017language} along with Layer Normalization (LN) \cite{ba2016layer} for the output of multi-head attention mechanism ${\alpha}^{\textit{safety }}$ to efficiently manage features. Formally,
\begin{flalign}
 {\bar{O}}^{t-t_{h}:t} &=  { \phi_{\textit {LN}}}\left(  { \phi_{\textit {MLP}}}( { \phi_{\textit {GLUs}}}({{\alpha}}))\right)
\end{flalign}
In particular, GLUs provide a mechanism to control the flow of information through the network, making the model more adaptable, which can be defined as:
\begin{equation}
    \phi_{\textit{GLUs}}(\alpha) = ( \alpha W_1 + b_1 ) \odot \phi_{\textit{sigmoid}}( \alpha W_2 + b_2 )
\end{equation}
where \( \alpha \) represents the safe attention coefficient from the multi-head attention mechanism, \( W_1 \) and \( W_2 \) are the learnable weight parameters associated with the GLUs layer, \( b_1 \) and \( b_2 \) are the corresponding biases, \( \odot \) denotes element-wise multiplication, \( \phi_{\textit{sigmoid}} \) is the sigmoid activation function, and \( \phi_{\textit{LN}}(\cdot) \) stands for Layer Normalization. Correspondingly, the output of the encoder within the Quantitative Safety Assessment is the high-level safety features, denoted as ${\bar{O}}^{t-t_{h}:t}_{\textit{safety}}$.

\subsubsection{Driver Behavior Profiling} \label{Behavior-aware Module_0}
{As shown in Figure \ref{fig1} (b), we represent vehicles and their interactions as nodes and edges, respectively, thereby constructing a Dynamic Geometric Graph (DGG). Leveraging this graph-based framework, we employ centrality measures from graph theory to profile continuum driver behavior in an unsupervised manner.

\textbf{Dynamic Geometric Graph.} Due to the dynamic nature of traffic scenarios, the structure of the DGG evolves over time. At any given moment \( t \), we define the DGG ${G}^{t} = \{V^{t},{E}^{t}\}$. Specifically, the node set \( {V^{t}} = \{{v_{0}^{t}},{v_{1}^{t}},...,{v_{n}^{t}}\} \), where node ${v_{i}^{t}}$ represents vehicle $i$. The adjacency matrix \( A^{t} \) illustrates whether edges exist between nodes, signifying the presence of interactions between vehicles. The establishment of this matrix is based on the distances between vehicles, which can be mathematically represented as follows:
\begin{equation}\label{eq.8}
    A^{t}(i, j)= \begin{cases}
    d(v_{i}^{t}, v_{j}^{t}) & \text{if } d(v_{i}^{t}, v_{j}^{t})\leq{r} \text{ and } i \neq j \\
    0 & \text{otherwise}
    \end{cases}
\end{equation}
{where $d(v_{i}^{t}, v_{j}^{t})$ denotes the distance between vehicle $i$ and vehicle $j$, and $r$ is a predefined threshold. The number of vehicles interacting with vehicle $i$ is represented as $\mathcal{N}_{i}^{t}$}

With these configurations in place, we then apply centrality measures to assess agent behavior, identify key agents, and evaluate the overall connectivity within the traffic graph.

\textbf{Centrality Measures.} Driver behavior significantly shapes the interaction patterns between the driver and surrounding agents, resulting in distinct spatiotemporal dynamics. Therefore, we posit that spatiotemporal dynamics can effectively differentiate between various driver behaviors. Given that centrality measures in graph theory provide a comprehensive description of the properties of nodes within a graph \cite{borgatti2018analyzing,zhang2017degree}, we employ centrality indicators such as degree $ {J}_{i}^{t}(D)$, closeness $ {J}_{i}^{t}(C)$, eigenvector ${J}_{i}^{t}(E)$, betweenness ${J}_{i}^{t}(B)$, power ${J}_{i}^{t}(P)$, and Katz ${J}_{i}^{t}(K)$ centrality to characterize the spatial interaction dynamics of agent $i$ at each moment $t$. To account for both the temporal and spatial dimensions of these dynamics, we further analyze the temporal evolution of these indicators and establish the Behavior-aware Criteria, enabling the continuous differentiation of diverse driving behaviors.

\textbf{1) Degree Centrality:} 
The number of agents a vehicle can influence reflects its significance within the traffic scene. Degree centrality $ {J}_{i}^{t}(D)$, a metric that measures the number of connections a node has, is thus naturally employed to describe the importance of each vehicle $i$. Formally,
\begin{equation}\label{eq.11_11}
\mathcal{J}_{i}^{t}(D)= \left| \mathcal{N}_{i}^{t}\right|+\mathcal{J}_{i}^{t-1}(D)
\end{equation}
where $\left| \mathcal{N}_{i}^{t}\right|$ denotes the total agents in $\mathcal{N}_{i}^{t}$.

\textbf{2) Closeness Centrality:} The position of a vehicle within a scene also reflects its significance. It is well-recognized that vehicles located centrally exert greater influence than those at the periphery. Consequently, closeness centrality $ {J}_{i}^{t}(C)$, which measures the proximity of a node to the center of the graph, is employed to characterize the importance of a vehicle as:
\begin{equation}\label{eq.14_21}
\mathcal{J}_{i}^{t}(C)=\frac{\left| \mathcal{N}_{i}^{t}\right|-1}{\sum_{\forall v_{j}^{t} \in \mathcal{N}^{t}_{i}}d\left(v_{i}^{t}, v_{j}^{t}\right)}
\end{equation}

\textbf{3) Eigenvector Centrality:}
The vehicle's behavior can influence a broader set of agents through those it directly interacts with, meaning that the importance of the directly connected agents also reflects the vehicle’s significance. Eigenvector centrality, which considers the importance of connected nodes, is used to assess the vehicle’s importance. The eigenvector centrality of the vehicle $i$ can be formulated as follows:
\begin{equation}\label{eq.13}
\mathcal{J}_{i}^{t}(E)=\frac{ \sum_{\forall v_{j}^{t} \in \mathcal{N}^{t}_{i}}d\left(v_{i}^{t}, v_{j}^{t}\right)}{\lambda}
\end{equation}
where $\lambda$ is the eigenvalue \cite{pillai2005perron}. 

\textbf{4) Betweenness Centrality:} The vehicle's influence can be transmitted to distant agents through intermediary agents, implying that vehicles frequently acting as intermediaries play a more crucial role in the network. Betweenness centrality $\mathcal{J}_{i}^{t}(B)$, a metric that measures the extent to which a node serves as an intermediary within the shortest path between any two nodes, is naturally used to assess the vehicle’s importance.
\begin{equation}\label{eq.14_2023}
\mathcal{J}_{i}^{t}(B) = \sum_{\forall v_{s}^{t},v_{k}^{t} \in {V}^{t}} \frac{\sigma_{j,k}(v_{i}^{t})}{\sigma_{j,k}}
\end{equation}
where $V^{t}$ denotes the set of all agents present in the scene, $\sigma_{j,k}$ signifies the total number of shortest paths between agent $v_{j}^{t}$ and agent $v_{k}^{t}$, and $\sigma_{j,k}(v_{i})$ represents the number of those paths traversing the agent $v_{i}^{t}$.

\textbf{5) Power Centrality:} {An interaction loop is a closed loop formed by a group of agents through direct or indirect interactions. A vehicle's participation in more interaction loops indicates greater influence within the overall traffic network. Power centrality $\mathcal{J}_{i}^{t}(P)$, which measures the frequency with which a node is part of closed cycles formed by edges, is used to describe the vehicle’s influence.}
\begin{equation}\label{eq.16-2023}
\mathcal{J}_{i}^{t}(P) = \sum_{k}\frac{A^{k}_{ii}}{k!}
\end{equation}
{where $A^{k}_{ii}$ denotes the $i$-th diagonal element of the adjacency matrix raised to the $k$-the power, $k!$ signifies the factorial.} 

\textbf{6) Katz Centrality:} 
{To address the limitation of degree centrality, which considers only direct interactions, we employ Katz centrality to emphasize both direct and distant interactions of the vehicle. Mathematically, the Katz centrality $\mathcal{J}_{i}^{t}(K)$ of an agent $v_{i}^{t}$ at time $t$ can be formulated as:} 
\begin{equation}\label{eq.15-2023}
\mathcal{J}_{i}^{t}(K) = \sum_{k} \sum_{j} \alpha^{k} A^{k}_{ij}+\beta^{k},  \forall i, j \in[0,n], \text { where } \alpha^{k} <\frac{1}{\lambda_{\max }}
\end{equation}
{where $n$ is the number of agents in the traffic scenario, $\alpha^{k}$ denotes the decay factor, $\beta^{k}$ represents weight for immediate neighbors, and $A^{k}_{ij}$ is the $i$,$j$-th element of the $k$-th power of the adjacency matrix. And $\lambda_{\max }$ denotes the largest eigenvalue of the adjacency matrix. By carefully selecting the value of the decay factor, Katz centrality can underscore the importance of closer interactions while discounting more distant connections.}

\textbf{Behavior-aware Criteria.} {Given the centrality metrics that capture the spatial interaction dynamics of traffic agents, we establish Behavior-aware Criteria that identify driving behavior not only based on the magnitude of these metrics but also on their temporal variation. Numerous studies have demonstrated the feasibility of this approach, showing that driving behavior can be identified using not only the instantaneous magnitude of features like speed but also their temporal derivatives, such as acceleration and jerk \cite{murphey2009driver}. This approach also aligns with human intuition, as driving behaviors characterized by large and fluctuating centrality measures over short periods are more likely to be relevant to driving behavior as sudden changes in acceleration within short intervals. Inspired by the established triadic relationship between velocity, acceleration, and jerk, we introduce three continuous criteria: Behavior Magnitude Index (BMI) ${\mathcal{C}}_{i}^{t}$, which measures the influence of driving behaviors by evaluating their centrality; Behavior Tendency Index (BTI) ${\mathcal{L}}_{i}^{t}$, which quantifies behavior propensity by calculating temporal derivatives, larger derivatives suggesting higher probabilities of specific behaviors; and Behavior Curvature Index (BCI) ${{\mathcal{I}}}_{i}^{t}$, which uses the jerk concept to measure the intensity of driving behaviors by calculating the second-order derivatives of continuous centrality measures. At time $t$, the behavior ${\mathcal{J}}_{i}^{{t}}$ for $v_{i}$ can be defined as ${\mathcal{J}}_{i}^{{t}}=\left[\mathcal{C}_{i}^{t},\mathcal{L}_{i}^{t},\mathcal{I}_{i}^{t}\right]^{T}$. {Each component meticulously evaluates the magnitude, probability, and intensity of diverse driving behaviors exhibited by the target vehicle and its surrounding agents. This assessment is conducted through the computation of threshold rates, gradients, and concavities associated with centrality measures, which capture behaviors such as lane changes, acceleration, and deceleration, as well as aggressive, neutral, or conservative driving tendencies. The underlying rationale is that driving behaviors characterized by substantial and volatile centrality measure values over short time intervals are more likely to exert a significant influence on nearby agents, emphasizing the temporal dynamics that are integral to human drivers' decision-making processes.}

\textbf{1) Behavior Magnitude Index.}
{The BMI is designed to quantify the scale and interconnectedness of various driving behaviors by assessing their centrality measures. The BMI encapsulates the absolute values of these measures, providing a quantitative representation of a behavior’s influence on the surrounding traffic agents. Specifically, the BMI focuses on each agent’s centrality measures, with a higher index indicating that a particular driving behavior exerts a more significant impact on the current traffic dynamics. Formally, we first formulate the BMI ${\mathcal{C}}$ for vehicle $i$ as follows:}

\vspace{-10pt}
{\begin{small}
\begin{equation}\label{eq.14}
 {\mathcal{C}}_{i}^{t}=\left[\left|{ \mathcal{J}^{t}_{i}(D)}\right|,\left|{ \mathcal{J}^{t}_{i}(C)}\right|,\left|{ \mathcal{J}^{t}_{i}(E)}
\right|,
\left|{ \mathcal{J}_{i}^{t}(B)}\right|, \left|{ \mathcal{J}_{i}^{t}(P)}\right|, \left|{ \mathcal{J}_{i}^{t}(K)}\right|\right]^{T}
\end{equation}
\end{small}}
where $\left| \cdot  \right|$  denotes the absolute value operator.

\textbf{2) Behavior Tendency Index.} 
{Building on the BMI, the BTI incorporates human factors in driving, particularly those behaviors that may cause significant fluctuations in centrality measures, such as aggressive driving, sudden lane changes, or abrupt braking. Specifically, the BTI aims to quantify the propensity for various driving behaviors by calculating their temporal derivatives. By capturing the temporal interaction dynamics of driving behavior, the BTI can identify instances where large gradients and local extrema suggest a higher probability of particular behaviors. This approach enables the model to estimate the likelihood of specific behaviors, even in the absence of explicit behavior classification. Mathematically,}

\vspace{-10pt}
{\begin{small}
\begin{flalign}\label{eq.14_1}
{\mathcal{L}}_{i}^{t}=\left|\frac{\partial\mathcal{\operatorname{{\mathcal{ C}}}}_{i}^{t}}{\partial t}\right| =\left[\left|\frac{\partial \mathcal{J}^{t}_{i}(D)}{\partial t}\right|,\left|\frac{\partial \mathcal{J}^{t}_{i}(C)}{\partial t }\right|,\cdots, \left|\frac{\partial \mathcal{J}_{i}^{t}(K)}{\partial t }\right|\right]^{T}
\end{flalign}
\end{small}}

\textbf{3) Behavior Curvature Index.}
{BCI introduces the concept of jerk to quantify the potential impact of driving behavior on surrounding agents. Building upon the BTI, the BCI captures driving behavior by calculating the second derivative of sequential centrality measures. The motivation behind BCI stems from the observation that abrupt changes in BTI over short periods, such as during braking or acceleration, result in peaks in the BCI curve. Additionally, BCI considers the duration of behavior fluctuations, positing that behaviors with prolonged fluctuations have a greater impact on the traffic environment than short-term variations. For instance, a driver who frequently changes lanes or adjusts speed over an extended period may confuse and pressure other drivers, thereby significantly disrupting the dynamic traffic environment.}

\vspace{-10pt}
{\begin{footnotesize}
\begin{flalign}\label{eq.15}
     {{\mathcal{I}}}_{i}^{t}  = \left|\frac{\partial\mathcal{\operatorname{{\mathcal{L}}}}_{i}^{t}}{\partial t}\right|
      = \left[\left|\frac{\partial^{2} \mathcal{J}_{i}^{t}(D)}{\partial^{2} t}\right|,\left|\frac{\partial^{2} \mathcal{J}_{i}^{t}(C)}{\partial^{2} t }\right|, \cdots, \left|\frac{\partial \mathcal{J}_{i}^{t}(K)}{\partial^{2} t }\right|\right]^{T} 
\end{flalign}
\end{footnotesize}}

{The introduction of the BMI, BTI, and BCI provides a holistic understanding of individual driving behaviors. Additionally, our proposed behavior-aware criteria eliminate the need for manual labeling during the training phase, effectively mitigating challenges associated with dynamic behavior labels and the selection of appropriate time windows.}

\textbf{Behavior Encoder.} To leverage the high-level prior knowledge embedded in the Behavior-aware Criteria, we introduce the behavior encoder to extract behavior features. The Behavior encoder comprises two main components: the LSTM and the multi-head self-attention mechanism. The behavior ${\mathcal{J}}_{i}^{t-t_{h}:t}$ are first processed by the LSTM, yielding temporal vectors:

\vspace{-10pt}
{\begin{small}
\begin{equation}
 {\tilde{J}}_{i}^{t-t_{h}:t}=\phi_{ \textit{LSTM}}\left({{h}}^{t-t_{h}:t}_{i},\phi_{\textit{MLP}}( {J}_{i}^{t-t_{h}:t}), \phi_{\textit{MLP}}( {\bar{O}_{\textit{safety}}}^{t-t{h}:t})\right)
\end{equation}
\end{small}}

{Here, the LSTM incrementally updates the hidden state of agent \( v_i \) on a frame-by-frame basis using shared weights. To enhance this process, we incorporate a multi-head self-attention mechanism and GLUs to calculate attention weights across diverse agent behaviors. This approach yields precise sequential behavioral features $\bar{O}_{\textit{behavior}}$, akin to the QSA:}
\begin{equation}
 {\bar{O}_{\textit{behavior}}}^{t-t_{h}:t} = \phi_{\textit {LN}}\left( \phi_{\textit {MLP}}(\phi_{\textit {GLUs}}({\alpha}^{\textit{behavior}}))\right)
\end{equation}
{where ${\alpha}^{\textit{behavior}}$ is the output of the multi-head self-attention mechanism within the behavior encoder.} }

\subsection{Priority-Aware Module}\label{Priority-Aware Pooling Module_0}
\subsubsection{Pooling Mechanism}
In light of recent advances in cognitive studies \cite{broadbent2023cognitive,liao2024bat}, it has become evident that the spatial positioning of vehicles within a scene can variably influence the behavior and decisions of a target vehicle. For instance, vehicles located directly in the anticipated trajectory path tend to exert greater influence relative to those situated behind. Furthermore, during overtaking maneuvers, vehicles positioned on the left may carry augmented significance. Recognizing these spatial intricacies, we introduce the Priority-Aware Module. This sophisticated module adeptly transforms the spatial coordinates of agents, encoding them into high-dimensional positional vectors, producing positional features.

Our pooling mechanism adeptly amalgamates dynamic positional information from the encompassing traffic scenario, effectively capturing both individual and multi-agent position vectors. This mechanism emphasizes the dynamic nuances of position data, accommodating historical agent states, denoted as \( \bm{S}_{i}^{t_{k}} \), as well as the intricate spatial interplay symbolized by \( \bm{P}_{i, j}^{t_{k}} \). Mathematically, these relationships are represented as:
\begin{equation}
\bm{S}_{i}^{t_{k}} = \{ p_{i}^{t_{k}} - p_{i}^{t_{k}-1}, v_{i}^{t_{k}} - v_{i}^{t_{k}-1}, a_{i}^{t_{k}} - a_{i}^{t_{k}-1} \}
\end{equation}
Correspondingly,
\begin{equation}
\bm{P}_{i, j}^{t_{k}} = \{ p_{i}^{t_{k}} - p_{j}^{t_{k}}, v_{i}^{t_{k}} - v_{j}^{t_{k}}, a_{i}^{t_{k}} - a_{j}^{t_{k}} \}
\end{equation}
By converting the position point sets into sequential vectors, this pooling mechanism effectively indicates the potential spatial relationship between the target agent and its neighboring agents at each scene and time step. This allows the module to accurately represent the necessary interactions over dynamic and spatial position characteristics of each agent.

\subsubsection{Priority Encoder} Within this module, the encoder, which combines both LSTM and multi-head attention mechanisms, processes the dynamic position vectors. This processing involves transforming discrete position vectors into a more continuous spatio-temporal domain, thereby improving the representation of temporal and spatial dynamics. At each discrete temporal instance \( t \), the encoder assimilates recent historical position vectors via an LSTM network:
\begin{equation}
 {{O}}_{\textit{priority}}^{t-t_{h}:t}=\phi_{ \textit{LSTM}}\left(\bm{\bar{h}}^{t-t_{h}:t-1}_{i}, \bm{S}^{t-t_{h}:t-1}_{i}, \bm{{P}}^{t-t_{h}:t-1}_{i,j}\right)
\end{equation}

Then, the output of the LSTM is then channeled through a multi-head attention mechanism and GLUs, similar to the quantitative safety assessment, culminating in the synthesis of refined priority features:
\begin{equation}
 {\bar{O}_{\textit{priority}}}^{t-t_{h}:t} =  { \phi_{\textit {LN}}}\left(  { \phi_{\textit {MLP}}}( { \phi_{\textit {GLUs}}}({\alpha}^{\textit{ priority}}))\right)
\end{equation}
where ${\alpha}^{\textit{ priority}}$ is the output of the multi-head attention mechanism in the priority encoder.

\begin{figure*}[t]
  \centering
\includegraphics[width=0.7\linewidth]{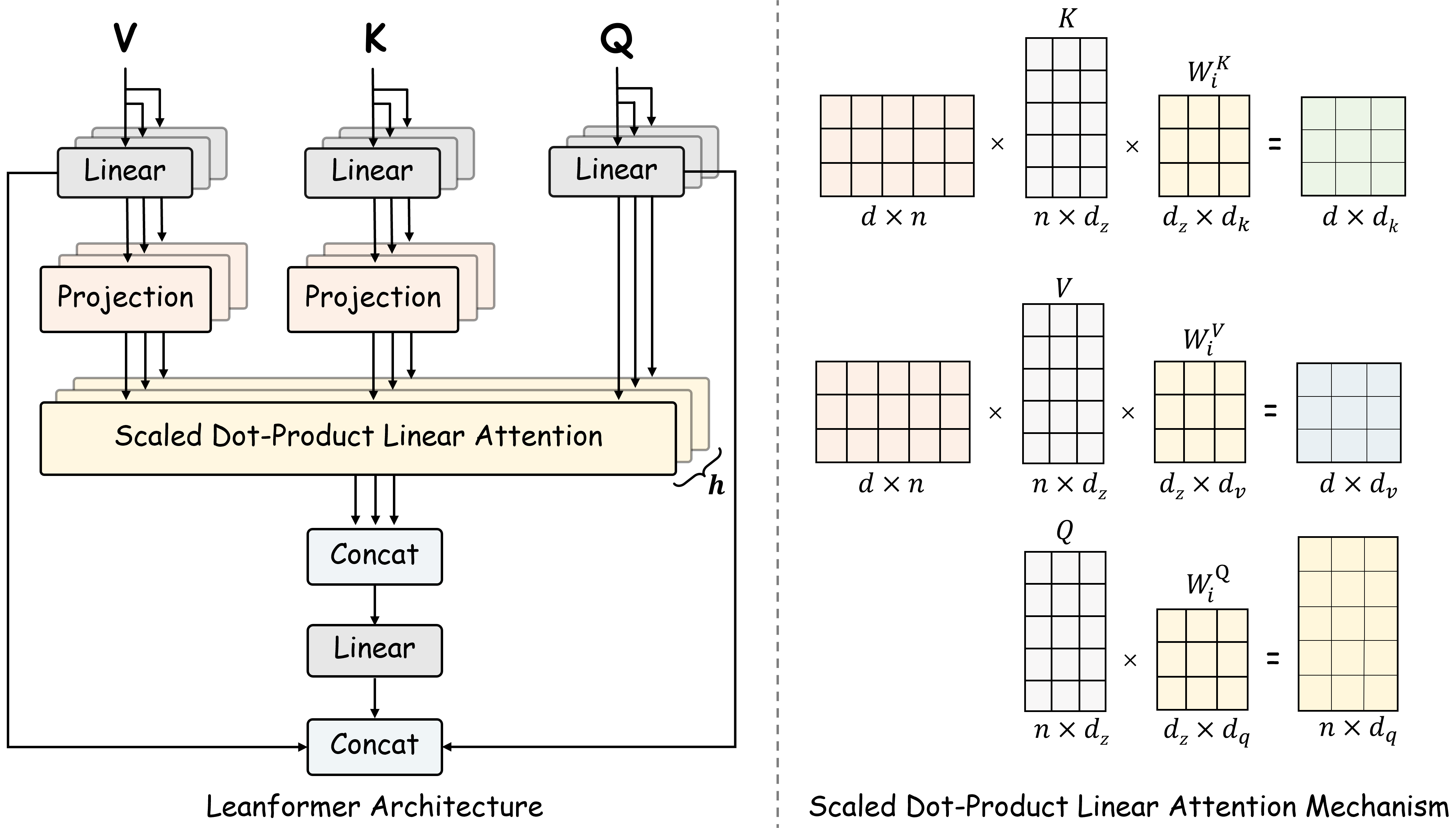} 
  \caption{{Architecture of the proposed Leanformer and the scaled dot-product linear attention mechanism.}}
  \label{fig7} 
\end{figure*}

\subsection{Interaction-Aware Module}\label{Interaction-Aware Pooling Module}
{
To better understand the synergistic influence of surrounding vehicles' risk levels, positions, and one's behavior on the target vehicle's future trajectory, we introduce an Interaction-Aware Module. This Module is based on a novel lightweight Transformer framework, i.e. Leanformer, which is an adaptation of the Linformer architecture \cite{wang2020linformer}. Departing from traditional Transformer models that rely on fully connected weight matrices, our framework refines the Transformer design, especially in its attention mechanism. By adopting a low-rank matrix approximation, we significantly increase the computational efficiency. This approach effectively reduces the computational complexity of self-attention from $O(n^2)$ to $O(n \times d)$, where $n$ represents the sequence length and $d$ denotes a much smaller projected dimension. Mathematically, the interaction is represented as follows:
\begin{equation}
     {{O}}^{t-t_{h}:t} = \phi_{\textit {MLP}}(\bar{O}_{\textit{safety}}^{t-t_{h}:t})\| \phi_{\textit {MLP}}({\bar{O}_{\textit{behavior}}}^{t-t_{h}:t})\| \phi_{\textit {MLP}}({\bar{O}_{\textit{ priority}}}^{t-t_{h}:t})
\end{equation}
This equation captures the integrated effect of safety perception, behavioral tendencies, and spatial positioning on the target vehicle's trajectory.

As shown in Figure \ref{fig7}, the sequence $ {{O}}^{t-t{h}:t}$ serves as the input to the Transformer-based framework. The queries ${Q}$, keys ${K}$, and values ${V}$ are obtained by performing linear transformations on the input sequence using low-rank projection matrices, as illustrated below:
\begin{equation}
\left\{\begin{array}{l}
{Q}={W}^{Q} (\phi_{\textit{MLP}} ({{O}}^{t-t_{h}:t}+L_{q}))\\
{K}={W}^{K} (\phi_{\textit{MLP}} ({{O}}^{t-t_{h}:t}+L_{k}))\\
{V}={W}^{V} (\phi_{\textit{MLP}}({{O}}^{t-t_{h}:t}))
\end{array}\right.
\end{equation}
Here, ${W}^{Q}$, ${W}^{K}$, and ${W}^{V}$ represent low-rank projection matrices for the queries, keys, and values, respectively. Notably, we introduce additional tokens $L_{q}$ and $L_{k}$ to the end of the queries $Q$ and keys $K$ to enhance feature representation and ensure training stability. This augmentation can be mathematically represented as follows:
\begin{equation}
\left\{\begin{array}{l}
L_{q}=\phi_{\textit{GRU}}\left(\phi_{\textit {MLP}}(\bar{O}_{\textit{safety}}^{t-t_{h}:t})\| \phi_{\textit {MLP}}({\bar{O}_{\textit{behavior}}}^{t-t_{h}:t})\right)
\\
L_{k}=\phi_{\textit{GRU}}\left(\phi_{\textit {MLP}}(\bar{O}_{\textit{safety}}^{t-t_{h}:t})\| \phi_{\textit {MLP}}({\bar{O}_{\textit{priority}}}^{t-t_{h}:t})\right)
\end{array}\right.
\end{equation}
where $\phi_{\textit{GRU}}$ denotes the GRU framework. The output matrix $ {\bar{O}}$ is computed as the sum of the outputs from all attention heads, denoted as $\overline{\textit{head}_i}$, where $i$ ranges from 1 to $h$, representing the total number of attention heads. Each attention head has its own set of projection matrices $W^{Q}_{i}$, $W^{K}_{i}$, and $W^{V}_{i}$, while skip connections are also used in this framework. This is expressed mathematically as follows:
\begin{equation}
 {\bar{O}} = \sum_{i=1}^{h} \overline{\textit{head}_i}+\left(\phi_{\textit {MLP}}(Q)\| \phi_{\textit {MLP}}(V)\right)
\end{equation}
In this equation, $\overline{\text{head}_i}$ refers to the output of the $i$-th attention head. The query matrix ${Q} \in {R}^{n \times d_{z}}$, key matrix ${K} \in \mathbb{R}^{n \times d_{z}}$, and value matrix ${V} \in \mathbb{R}^{n \times d_{z}}$ are all of dimensionality $n \times d_{z}$.

The attention mechanism in each attention head computes a context mapping matrix $\bar{{P}}$ of size $(k \times d)$ using scaled dot-product attention. It involves the following calculations:
\begin{equation}
\bar{{P}} = \phi_{\text{softmax}}\left(\frac{{Q} {W}^{Q}_{i}\left({U}_{i} {K} {W}^{K}_{i}\right)^T}{\sqrt{d_k}}\right)
\end{equation}
where ${U}_{i} \in \mathbb{R}^{n \times k}$ denotes a fixed linear projection matrix, while $d_{k}$ is the dimensionality of the projected key vectors.

Finally, the output matrix $\overline{\text{head}_i}$ is obtained by multiplying $\bar{{P}}$ with the projected value matrix ${F} {V} {W}^{V}_{i}$ using a linear projection matrix ${F}_{i} \in \mathbb{R}^{n \times k}$:
\begin{equation}
\overline{\text{head}_i} = \bar{{P}} {F} {V} {W}^{V}_{i} =  {\phi_{\textit{Attention}}}\left(Q W_{i}^{Q}, U_{i} K W^{K}_{i}, {F}_{i} V W_{i}^{V}\right)
\end{equation}
This can also be expressed as follows:
\begin{equation}
\begin{aligned}
\overline{\textit{head}_{i}}  
=\underbrace{ {\phi_{\text{softmax}}}\left(\frac{Q W_{i}^{Q} (U_{i} K W_{i}^{K})^{T}}{\sqrt{d_{k}}}\right)} _{\bar{P}: n \times k} \cdot \underbrace{{F}_{i} V W_{i}^{V}}_{k \times d}
\end{aligned}
\end{equation}
where the attention weights are obtained by calculating the scaled dot product of the query and key projection matrices, followed by the softmax activation function. The resulting weights are then used to compute the weighted sum of the value projection matrix, which represents the composite interactive vectors $\bar{O}$ fed into the Multimodal Decoder to generate the future trajectories for the target vehicle.}

\subsection{Multimodal Decoder}
The decoder, rooted in a Gaussian Mixture Model with multimodality, employs a dedicated LSTM and a fully connected layer. It processes the composite interactive vectors \( {\bar{O}} \) to forecast the target vehicle's trajectory. The predicted trajectory, \( \boldsymbol{\bm{Y}_{0}^{t:t+t_{f}}} \), is determined by:
\begin{equation}
    \boldsymbol{\bm{Y}_{0}^{t:t+t_{f}}} = {F}_{\theta}\left( {F}_{\theta}( {\bar{O}})\right)
\end{equation}
such that,
\begin{equation}
      {F}_{\theta}(\bm{\cdot})=\phi_{\textit{ReLU}}\left(\phi_\textit{MLP}\left[\phi_{\text {GN}}\left(\phi_{\textit {LSTM}}(\bm{\cdot})\right)\right]\right)
\end{equation}

Here, \( \phi_{\text{GN}} \) is Group Normalization, used for improved training stability. The decoder's output comprises multiple future trajectories for the vehicle.

\section{Experiments}\label{Experiments}

\subsection{Experimental Setups}
{
To validate the prediction capability of our model across different scenarios, we conduct a series of experiments on three widely used traffic datasets: NGSIM, MoCAD, and HighD. The experiments on NGSIM and HighD primarily assess the model's performance in highway scenarios with varying traffic densities, while the experiments on MoCAD focus on its ability to predict in urban and campus-like, unstructured environments.
To ensure a fair comparison with existing models \cite{wang2023wsip, gao2023dual}, we adopt the same training and evaluation protocols as those used in prior benchmarks. Specifically, we define a reference time point, utilizing the preceding three seconds of data as input to the model and the subsequent five seconds as ground truth for supervising model training. In addition, we provide a detailed analysis of our model's performance across both \textbf{short-term} ($\leq2$ seconds) and \textbf{long-term} prediction horizons ($>$2 seconds).  

Acknowledging a gap in existing research regarding data omissions in prediction, we develop an innovative approach to tackle the issue of missing data. We establish the \textit{missing} test set, which is further categorized into three subsets based on the duration of data omissions: \textit{drop 3-frames}, \textit{drop 5-frames}, and \textit{drop 8-frames}. Omissions are purposefully made around the midpoint of the historical trajectory. For instance, in the \textit{drop 5-frames} subset, data ranging from the $(t-8)$-th to the $(t-12)$-th frame is excluded. To manage these omissions, we employ simple linear interpolation. The evaluation results on these subsets are reported as \textbf{CITF (drop 3-frames)}, \textbf{CITF (drop 5-frames)}, and \textbf{CITF (drop 8-frames)}, respectively. Furthermore, to demonstrate the adaptability and efficiency of our model, we train it on a \textit{limited} training set, which contains only 25\% of the available training datasets. This model variant is denoted as \textbf{CITF (25\%)} in this study.}

\subsection{Evaluation Metric}
To evaluate the performance of our model and compare it to the baselines, we utilize the commonly used performance metric: Root Mean Square Error (RMSE), which can be defined as follows:
\begin{equation}
\textit{RMSE} = \sqrt{\frac{1}{n} \sum_{i=1}^{n} \left(\mathbf{Y}^{\text{pred}}(T) - \mathbf{Y}^{\text{gt}}(T)\right)^2}
\end{equation}
where $\mathbf{Y}^{\text{pred}}(T)$ and $\mathbf{Y}^{\text{gt}}(T)$ represent the predicted and ground-truth trajectories, respectively, at a given horizon $T$.

\subsection{Training and Implementation Details}\label{Training} 
{Our model was trained on an NVIDIA A40 GPU with 48GB of memory. The training process utilized a batch size of 64 and was conducted over 20 epochs. We adopted a dynamic learning rate strategy, initially set at $10^{-3}$ and gradually reducing to $10^{-5}$. The Adam optimizer was employed, coupled with the CosineAnnealingWarmRestarts scheduler to manage the learning rate adjustments. Following the multi-task learning framework \cite{kendall2018multi}, our loss function combines the RMSE and the Negative Log-Likelihood (NLL) metrics. The RMSE metric quantifies the average Euclidean distance between predicted and ground truth trajectories, serving as a general measure of predictive accuracy. The NLL metric is particularly valuable for assessing the fidelity of trajectory predictions relative to expected maneuvers, ensuring the reliability of trajectory forecasting within action-based models. }

\begin{table*}[htbp]
  \centering
  \caption{{Evaluation results of the proposed model and baseline methods on the \textbf{short-term} prediction horizon ($\leq2$ seconds). The metric is RMSE (m), with lower values indicating better performance. Cells marked with (`-') indicate data not available. \textbf{Bold} and 
 \underline{underlined} values indicate the best and the second-best performance in each category, respectively.}}  \setlength{\tabcolsep}{9mm}
  \resizebox{0.85\linewidth}{!}{
    \begin{tabular}{c|cc|cc|cc}
    \toprule
    \multicolumn{1}{c}{\multirow{2}[3]{*}{Model}} & \multicolumn{2}{c}{NGSIM} & \multicolumn{2}{c}{MoCAD} & \multicolumn{2}{c}{HighD}\\
\cmidrule{2-7}   \multicolumn{1}{c}{}   & 1     & 2         & 1     & 2      & 1     & 2          \\
    \hline
           CS-LSTM \cite{deo2018convolutional} & 0.61  & 1.27    & 1.45  & 1.98    & 0.22  & 0.61  \\
           NLS-LSTM \cite{messaoud2019non} & 0.56  & 1.22    & 0.96  & 1.27   & 0.20  & 0.57  \\
           CF-LSTM \cite{xie2021congestion} & 0.55  & 1.10    & 0.72  & 0.91   & 0.18  & 0.42   \\
            iNATran \cite{chen2022vehicle} & {0.39}  &0.96    & -  & -  & \textbf{0.04}  & \textbf{0.05}  \\
            BAT \cite{liao2024bat} & \textbf{0.23} & \textbf{0.81}  & \underline{0.35}  & \underline{0.74}  & 0.08 & 0.14  \\
           MHA-LSTM \cite{messaoud2021attention} & 0.41  & 1.01  & 1.25  & 1.48  & 0.19  & 0.55 \\
           STDAN \cite{chen2022intention} & {0.39}  & 0.96    & 0.62  & 0.85   & 0.19  & 0.27 \\
           HLTP  \cite{10468619} & {0.41}    & {0.91}  & {0.55} &{0.76} &{0.09} & {0.16}\\
            HLTP++ \cite{liao2024less} & 0.46 & {0.98} & {0.64} & {0.86} & {0.12}  &{0.18}\\
            WSiP \cite{wang2023wsip} & 0.56  & 1.23   & 0.70  & 0.87   & 0.20  & 0.60 \\
    
    \hline
\textbf{CITF} & \underline{0.30} & \textbf{0.81} & \textbf{0.28} & \textbf{0.63} & \textbf{0.04} & \underline{0.09} \\
   \textbf{CITF (drop 3-frames)} & {0.38} & \underline{0.86} & \underline{0.35} & {0.80} & \underline{0.05} & {0.11} \\
   \textbf{CITF (drop 5-frames)} & {0.41} & {0.90} & {0.45} & {0.94} & {0.16} & {0.30} \\
   \textbf{CITF (drop 8-frames)} & {0.42} & {0.94} & {0.65} & {1.03} & {0.17} & {0.44} \\
   \textbf{CITF (25\%)} & {0.42} & {0.93} & {0.55} & {0.96} & {0.08} & {0.21} \\
    \bottomrule
    \end{tabular}%
    }
  \label{short_term}%
\end{table*}%

\begin{table*}[htbp]
  \centering
  \caption{{Comparative evaluation of our model on the \textbf{short-term} prediction horizon against selected baselines on the \textit{missing} test set of the NGSIM dataset. RMSE (m) is used as the evaluation metric. \textbf{Bold} values indicate the best performance, while \underline{underlined} values indicate the second-best performance in each category.}}
  \setlength{\tabcolsep}{10mm}
  \resizebox{0.85\linewidth}{!}{
    \begin{tabular}{c|cc|cc|cc}
    \toprule 
     \multicolumn{1}{c}{\multirow{2}[3]{*}{Model}}  & \multicolumn{2}{c}{\makecell{Drop 3-frames}} & \multicolumn{2}{c}{\makecell{Drop 5-frames}} & \multicolumn{2}{c}{\makecell{Drop 8-frames}} \\
    \cmidrule{2-7} \multicolumn{1}{c}{}
    & 1s & 2s & 1s & 2s & 1s & 2s \\
\hline
    \makecell{CS-LSTM \cite{deo2018convolutional}} & 0.67 & 1.47 & 0.75 & 1.52 & 0.84 & 1.72 \\
    \makecell{CF-LSTM \cite{xie2021congestion}} & 0.59 & 1.14 & 0.64 & 1.37 & 0.70 & 1.46 \\
    \makecell{WSiP \cite{wang2023wsip}} & 0.60 & 1.29 & 0.69 & 1.37 & 0.76 & 1.58 \\
    \makecell{STDAN \cite{chen2022intention}} & 0.42 & 1.00 & \underline{0.47} & 1.12 & 0.57 & 1.37 \\
    \makecell{BAT \cite{liao2024bat}} & \textbf{0.28} & \underline{0.88} & 0.48 & \underline{0.99} & \underline{0.52} & \underline{1.15} \\

 \makecell{HLTP  \cite{10468619}} & {0.49}    & {1.21}  & {0.67} &{1.34} &{0.75} & {1.46}\\
\makecell{ HLTP++ \cite{liao2024less}} & 0.48 & {1.09} & {0.63} & {1.26} & {0.70}  &{1.34}\\
    \hline
 \textbf{CITF} & \underline{0.38} & \textbf{0.86} & \textbf{0.41} & \textbf{0.90} & \textbf{0.42} & \textbf{0.94} \\
    \bottomrule
    \end{tabular}%
  }
  \label{NGSIM_missing_vertical_short}%
\end{table*}%

\subsection{Experiment Results}{
\subsubsection{Performance Comparison on \textbf{Short-term} Horizon}
We present comparative results of our model's prediction performance on the \textbf{short-term} prediction horizon ($\leq2$ seconds) against existing baseline models on the NGSIM, MoCAD, and HighD datasets in Table \ref{short_term}. Our model achieves the best or second-best performance across all three datasets. Among existing approaches, the BAT model demonstrates the best results on the NGSIM and MoCAD datasets, while the iNATran model performs best on the HighD dataset. Therefore, a comparative analysis of these two models is both essential and representative. Compared to the BAT model, although our model slightly underperforms in the 1-second prediction horizon on the NGSIM dataset, it surpasses BAT on all other datasets. Specifically, on the MoCAD dataset, our model achieves a minimum improvement of 14.8\% in short-term prediction horizons, while on the HighD dataset, the improvement is at least 35.8\%. Similarly, when compared to the iNATran model, our model falls slightly behind only in the 2-second prediction horizon on the HighD dataset. However, it demonstrates significant improvements on the NGSIM dataset, achieving gains of 23.1\% and 15.6\% in the 1-second and 2-second prediction horizons, respectively. We further conducted a comprehensive evaluation of the robustness of our proposed model using the missing test set and compared its performance with SOTA baselines. As shown in Table \ref{short_term}, CITF outperforms most models, including CS-LSTM and WSiP, on the \textit{missing} test set, even surpassing their performance on the complete test set. Additionally, we assessed the ability of various baselines to handle data missingness challenges using the NGSIM \textit{missing} dataset, with results presented in Table \ref{NGSIM_missing_vertical_short}. A clear trend emerges: as the proportion of missing data increases, the advantages of CITF become even more pronounced. Specifically, CITF outperforms the previous best-performing BAT model by at least 9.0\% and 20.0\% in the \textit{drop 5-frames} and \textit{drop 8-frames} scenarios, respectively. Overall, while CITF does not show a distinct advantage over previous baseline models in short-term prediction horizons, it significantly outperforms them in the presence of data omissions—a common challenge in real-world applications due to observational constraints.

\subsubsection{Performance Comparison on \textbf{Long-term} Horizon}
As shown in Table \ref{long_term}, our model achieves the best performance across all datasets for \textbf{long-term} prediction horizons ($>$2 seconds). On the NGSIM dataset, our model significantly surpasses all baselines from 2018 to 2024, achieving improvements of 7.8\%, 15.7\%, and 17.1\% for the 3-second, 4-second, and 5-second prediction horizons, respectively. Similarly, results on the MoCAD dataset highlight the strong performance of our model in busy urban traffic scenarios, with improvements of 21.6\%, 23.7\%, and 20.8\% over the same horizons. Furthermore, our model demonstrates substantial performance gains on the HighD dataset, outperforming the BAT and DACR-AMTP models with remarkable improvements of up to 31.8\% and 57.4\% in RMSE, respectively. In the 5-second prediction horizon, our model surpasses iNATran by an impressive 60.9\%. 
We also investigated the impact of data omissions on the model's long-term prediction performance. As shown in Table \ref{long_term}, even when faced with the \textit{drop 3-frames} and \textit{drop 5-frames} scenarios, CITF outperforms existing baselines on the majority of prediction horizons across all datasets. Specifically, at the 5-second prediction horizon, CITF (drop 3-frames) achieved improvements of 13.2\%, 16.7\%, and 24.2\% on the NGSIM, MoCAD, and HighD datasets, respectively. Table \ref{NGSIM_missing_vertical_long} further illustrates CITF's performance against other models on the missing test set of the NGSIM dataset. CITF outperforms all other models across all prediction horizons, and this advantage increases as the proportion of data omissions grows. On the \textit{drop 8-frames} test set, CITF surpasses the best existing models by 9.3\%, 8.9\%, and 11.8\% at the 3, 4, and 5-second prediction horizons, respectively. Overall, CITF exhibits better performance in long-term prediction horizons compared to short-term horizons. These results underscore the model’s ability to capture long-term intentions and deliver accurate predictions over long-term horizons.

}

\begin{table*}[htbp]
  \centering
  \caption{{Evaluation results of the proposed model and baseline methods on the \textbf{long-term} prediction horizon ($>$2 seconds). The accuracy is measured using RMSE (m), with lower values indicating better performance. Cells marked with (`-') indicate data not available. \underline{underlined} values indicate the best and the second-best performance in each category, respectively.
  }}  \setlength{\tabcolsep}{6mm}
  \resizebox{0.85\linewidth}{!}{
    \begin{tabular}{c|ccc|ccc|ccc}
    \toprule
    \multicolumn{1}{c}{\multirow{2}[3]{*}{Model}} & \multicolumn{3}{c}{NGSIM} & \multicolumn{3}{c}{MoCAD} & \multicolumn{3}{c}{HighD}\\
\cmidrule{2-10}   \multicolumn{1}{c}{}   & 3     & 4     & 5     & 3     & 4     & 5    & 3     & 4     & 5     \\
    \midrule
           CS-LSTM \cite{deo2018convolutional} & 2.09 & 3.10  & 4.37   & 2.94& 3.56  & 4.49   & 1.24& 2.10  & 3.27\\
           NLS-LSTM \cite{messaoud2019non}  & 2.02 & 3.03  & 4.30    & 2.08  & 2.86  & 3.93 & 1.14  & 1.90  & 2.91 \\
           CF-LSTM \cite{xie2021congestion}  & 1.78  & 2.73  & 3.82   & 1.73& 2.59  & 3.44   & 1.07 & 1.72  & 2.44 \\
            iNATran \cite{chen2022vehicle}   &1.61 & 2.42  & 3.43 & -  & -  & -  & 0.21 & 0.54  & 1.10\\
            BAT \cite{liao2024bat} & 1.54 & 2.52  & 3.62  & {1.39} & {2.19} &{2.88}  & \underline{0.20} & 0.44  & 0.62\\
           MHA-LSTM \cite{messaoud2021attention} & 1.74 & 2.67  & 3.83    & 2.57 & 3.22  & 4.20   & 1.10 & 1.84  & 2.78 \\
           STDAN \cite{chen2022intention}  & 1.61& 2.56 & 3.67   & 1.62  & 2.51  & 3.32  & 0.48  & 0.91  & 1.66 \\
            WSiP \cite{wang2023wsip}  & 2.05 & 3.08  & 4.34     & 1.70 & 2.56  & 3.47  & 1.21 & 2.07  & 3.14\\
 HLTP++ \cite{liao2024less} & 1.52 & {2.17} & {3.02} & {1.56} & {2.40}  &{3.19} & {0.30} & {0.47}  &{0.75}\\
    
   \hline
    \textbf{CITF}  & \textbf{1.42}  & \textbf{2.04} & \textbf{2.82}   & \textbf{1.09} & \textbf{1.67} & \textbf{2.28}  & \textbf{0.18} & \textbf{0.30} & \textbf{0.43}  \\
    \textbf{CITF (drop 3-frames)}  & \underline{1.51}  & \underline{2.32} & \underline{2.95}   & \underline{1.24} & \underline{1.75} & \underline{2.40}  & 0.23 & \underline{0.35} & \underline{0.47}  \\
    \textbf{CITF (drop 5-frames)}  & 1.52  & 2.40 & 3.31   & {1.30} & 1.80 & 2.72  & 0.43 & 0.64 & 0.92  \\
    \textbf{CITF (drop 8-frames)}  & 1.65  & 2.45 & 3.51   & {1.63} & 2.13 & 2.98  & 0.83 & 1.25 & 1.72  \\
    \textbf{CITF (25\%)}  & 1.55  & 2.54 & 3.30   & {1.35} & 2.22 & 3.10  & 0.41 & 0.62 & 0.92  \\
    \bottomrule
    \end{tabular}%
    }
  \label{long_term}%
\end{table*}%

\begin{table*}[htbp]
  \centering
  \caption{{Comparative evaluation of our model on the \textbf{long-term} prediction horizon against selected baselines on the \textit{missing} test set of the NGSIM dataset. RMSE (m) is used as the evaluation metric. \textbf{Bold} values indicate the best performance, while \underline{underlined} values indicate the second-best performance in each category.}}
  \setlength{\tabcolsep}{5mm}
  \resizebox{0.8\linewidth}{!}{
    \begin{tabular}{c|ccc|ccc|ccc}
    \toprule
    \multicolumn{1}{c}{\multirow{2}[3]{*}{Model}} & \multicolumn{3}{c}{\makecell{Drop 3-frames}} & \multicolumn{3}{c}{\makecell{Drop 5-frames}} & \multicolumn{3}{c}{\makecell{Drop 8-frames}} \\
    \cmidrule{2-10} \multicolumn{1}{c}{}
    & 3s & 4s & 5s & 3s & 4s & 5s & 3s & 4s & 5s \\
    \hline
    \makecell{CS-LSTM \cite{deo2018convolutional}} & 2.34 & 3.60 & 4.71 & 2.47 & 3.82 & 4.97 & 2.64 & 3.97 & 5.34 \\
    \makecell{CF-LSTM \cite{xie2021congestion}} & 1.82 & 2.77 & 3.91 & 1.94 & 2.83 & 3.98 & 2.21 & 3.10 & 4.47 \\
    \makecell{WSiP \cite{wang2023wsip}} & 2.10 & 3.17 & 4.42 & 2.19 & 3.41 & 4.77 & 2.37 & 3.64 & 5.10 \\
    \makecell{STDAN \cite{chen2022intention}} & 1.68 & 2.64 & 3.72 & 1.81 & 2.75 & 3.88 & 2.14 & 3.04 & 4.19 \\
    
    \makecell{BAT \cite{liao2024bat}} & \underline{1.59} & \underline{2.59} & \underline{3.67} & \underline{1.74} & \underline{2.61} & \underline{3.84} & \underline{1.82} & \underline{2.69} & \underline{3.98} \\
   \hline
   \textbf{CITF} & \textbf{1.51} & \textbf{2.32} & \textbf{2.95} & \textbf{1.52} & \textbf{2.40} & \textbf{3.31} & \textbf{1.65} & \textbf{2.45} & \textbf{3.51}\\
    \bottomrule
    \end{tabular}%
  }
  \label{NGSIM_missing_vertical_long}%
\end{table*}%

\subsubsection{Performance Comparison on \textbf{Limited} 25\% Training Set}
To challenge our model's adaptability, we trained it using only a quarter of the available training set from the NGSIM, HighD, and MoCAD datasets, yet evaluated its performance on the \textit{complete} test set. Impressively, as shown in Tables \ref{short_term} and \ref{long_term}, even with this limited training data, our model delivered RMSE values that were notably lower than most baseline models. Such results underscore our model's efficiency and robustness in trajectory prediction. This performance indicates a promising potential: our model might substantially cut down on the data demands typically associated with training AVs, particularly in scenarios that are data-scarce. In summary, our findings attest to the model's reliability, resource efficiency, and precision in forecasting vehicle trajectories.

\subsubsection{Comparative Analysis of Model Performance and Complexity} 
As shown in Table \ref{tab:param}, our model is benchmarked against several top baselines across three real-world datasets. A notable challenge in this field is the limited availability of efficiency metrics, compounded by restricted access to the source code of various models. Consequently, our comparison mainly focuses on open-source models. Although our model is not the most parameter-efficient, it surpasses all competitors by achieving the lowest average RMSE across all datasets. Remarkably, this high level of accuracy is attained with substantially reduced model complexity—using 18.3\% fewer parameters than Gava. Additionally, we assess the inference speed of CITF on the NGSIM dataset. As presented in Table \ref{second}, while our model’s inference speed is slightly slower than that of the MHA-LSTM model, it ranks among the most accurate. Specifically, CITF outpaces the previous SOTA model STDAN, with inference speeds 31.8\% faster, respectively. These findings highlight CITF’s ability to balance speed and accuracy, further emphasizing its lightweight, efficient, and precise performance in predicting future vehicle trajectories.

\begin{table}[htbp]
  \centering
  \caption{Comparative evaluation of CITF with selected baselines. Highlighting the accuracy metric (Average RMSE (m)) and complexity measured by the number of parameters \#Param. (K). \textcolor{blue!50}{Purple} indicates the performance of our model. }
  \setlength{\tabcolsep}{2mm}
  \resizebox{0.8\linewidth}{!}{
    \begin{tabular}{ccccccc}
    \bottomrule
    \multirow{2}[4]{*}{Model} & \multicolumn{3}{c}{Average RMSE (m)} & \multirow{2}[2]{*}{\#Param. (K)}\\
\cmidrule{2-4}          & NGSIM     & HighD   &MoCAD   \\
    \midrule
    CS-LSTM \cite{deo2018convolutional} & 2.29 & 1.49 &2.88 & \textbf{194.92}\\
    CF-LSTM \cite{xie2021congestion}& 1.99 & 1.17 &1.88 & 387.10 \\
    WSiP \cite{wang2023wsip}& 2.25 & 1.44 &1.86 & 300.76\\
    GaVa \cite{liao2024human}& \underline{1.65} & \underline{0.39} & \underline{1.63} & 360.75 \\
   \rowcolor{blue!8} \textbf{CITF} & \textbf{1.48} & \textbf{0.21}  &\textbf{1.00} &  \underline{294.61}\\
   \toprule 
    \label{tab:param}
\end{tabular}}%
\end{table}%

\begin{table}[htbp]
    \centering
    \caption{Inference time comparison of CITF with the SOTA baselines on NGSIM. The inference time is for 10 batches with a size of 128 on two Nvidia A40 48G GPUs. \textcolor{blue!50}{Purple} indicates the performance of our model. }
      \resizebox{0.8\linewidth}{!}{
    \begin{tabular}{ccc}
     \bottomrule
        {Model} & {Average RMSE (m)} & {Inference time (s)} \\
        \hline
        CS-LSTM \cite{deo2018convolutional}& 2.29 & 0.22 \\
        MHA-LSTM \cite{messaoud2020attention}& 1.93 & \textbf{0.11} \\
        TS-GAN \cite{wang2020multi}& 2.06 & 0.23 \\
        WSiP \cite{wang2023wsip}& 2.25 & 0.25\\
        STDAN \cite{chen2022intention}& \underline{1.87} & 0.22 \\
        \rowcolor{blue!8}\textbf{CITF} & \textbf{1.48} & \underline{0.15} \\
        \toprule 
        \label{second}
    \end{tabular}
    }
\end{table}

\subsection{Ablation Studies} \label{Ablation Studies}
We perform a detailed ablation study to assess the specific contributions of each component within our trajectory prediction model.
The summarized results are presented in Table \ref{table_5}. Notably, Model F, which integrates all components, consistently outperforms other variations across all evaluation metrics, underscoring the combined value of these components in optimizing performance. In contrast, Model A, which omits the Driver Behavior Profiling within the Perceived Safety-Aware Module, experiences a substantial decline in performance, particularly in short-term predictions, with reductions of at least 19.8\% on the NGSIM dataset and 30.7\% on the HighD dataset.
This underscores the critical role of Driver Behavior Profiling in improving trajectory prediction accuracy.
Model B, a reduced version of Model F without the Perceived Safety-Aware Module shows a significant reduction in RMSE, especially for long-term predictions, with improvements of at least 16.0\% and 28.0\% on the NGSIM and HighD datasets, respectively. This highlights the importance of considering perceived safety factors in trajectory prediction, especially for long-term prediction.
Model C, which uses absolute coordinates instead of relative positions in the Priority-Aware Module, displays non-negligible reductions in prediction metrics, highlighting the importance of spatial relationships in achieving accuracy.
Model D, which lacks the Interaction-Aware Module, shows performance losses of at least 14.8\% and 25\% on the NGSIM and HighD datasets for short-term prediction, and at least 8.5\% and 26.7\% for long-term prediction, respectively. 
Finally, Model E reduces the multimodal probabilistic maneuver prediction in the Decoder, resulting in a performance degradation of at least 5.7\% and 13.9\% in the NGSIM and HighD datasets, respectively. This suggests their importance in improving prediction accuracy. 

\begin{table}[t]
  \centering
  \caption{Different components of ablation study.}\label{Tablem}
  \resizebox{0.8\linewidth}{!}{
    \begin{tabular}{ccccccc}
     \bottomrule
    \multirow{2}[4]{*}{Components} & \multicolumn{6}{c}{Ablation methods} \\
\cmidrule{2-7}          & A     & B     & C     & D     & E  &F\\
   \hline
    Driver Behavior Profiling & \ding{56} & \ding{52} & \ding{52} & \ding{52} & \ding{52} & \ding{52} \\
     Perceived Safety-Aware Module & \ding{52} & \ding{56} & \ding{52}& \ding{52}  & \ding{52} & \ding{52} \\
    Priority-Aware Module & \ding{52} & \ding{52} &  \ding{56} 
 &\ding{52} & \ding{52}  & \ding{52} \\
    Interaction-Aware Module & \ding{52} & \ding{52} & \ding{52} & \ding{56} & \ding{52}  & \ding{52} \\
     Multimodal Decoder & \ding{52} & \ding{52} & \ding{52} & \ding{52}& \ding{56}  & \ding{52} \\
   \toprule
    \end{tabular}}%
\end{table}%
\begin{table}[htb]
  \centering
  \caption{Evaluation results of the ablation analysis for different models on the NGSIM and HighD datasets, with RMSE (m) as the evaluation metric.}
  \setlength{\tabcolsep}{3mm}
  \resizebox{0.95\linewidth}{!}{
    \begin{tabular}{c|ccccccc}
    \bottomrule
    \multicolumn{1}{c}{\multirow{2}[4]{*}{Dataset}} & \multirow{2}[4]{*}{Time (s)} & \multicolumn{6}{c}{ Model} \\
\cmidrule{3-8}    \multicolumn{1}{c}{} &       & A     & B     & C     & D     &E & F \\
    \hline
    \multirow{5}[2]{*}{NGSIM} & 1     & 0.47 & 0.45  & 0.41  & 0.43 &0.39 & \textbf{0.30} \\
          & 2     & 1.01  & 0.97  & 0.89  & 0.93 &0.88 & \textbf{0.81} \\
          & 3  &1.50   & 1.69  & 1.77  & 1.54  & 1.59 & \textbf{1.42} \\
          & 4     & 2.58  & 2.70  & 2.39  & 2.47  &2.27 & \textbf{2.04} \\
          & 5     & 3.32 & 3.41  & 3.10  & 3.26  &2.98 & \textbf{2.82} \\
    \hline
    \multirow{5}[2]{*}{HighD} & 1     & 0.06  & 0.05  & 0.04  & 0.05 &0.05& \textbf{0.04} \\
          & 2     & 0.13  & 0.14  & 0.10 & 0.12 &0.11 & \textbf{0.09} \\
          & 3     & 0.25  & 0.25  & 0.21  & 0.24 & 0.22 & \textbf{0.18} \\
          & 4     & 0.40  & 0.42  & 0.36  & 0.38 & 0.36 & \textbf{0.30} \\
          & 5     & 0.58  & 0.67  & 0.51  & 0.55  &0.49 & \textbf{0.43} \\
   \toprule
    \end{tabular}%
    }
  \label{table_5}%
\end{table}%

\begin{figure}[t]
  \centering  \includegraphics[width=0.95\linewidth]{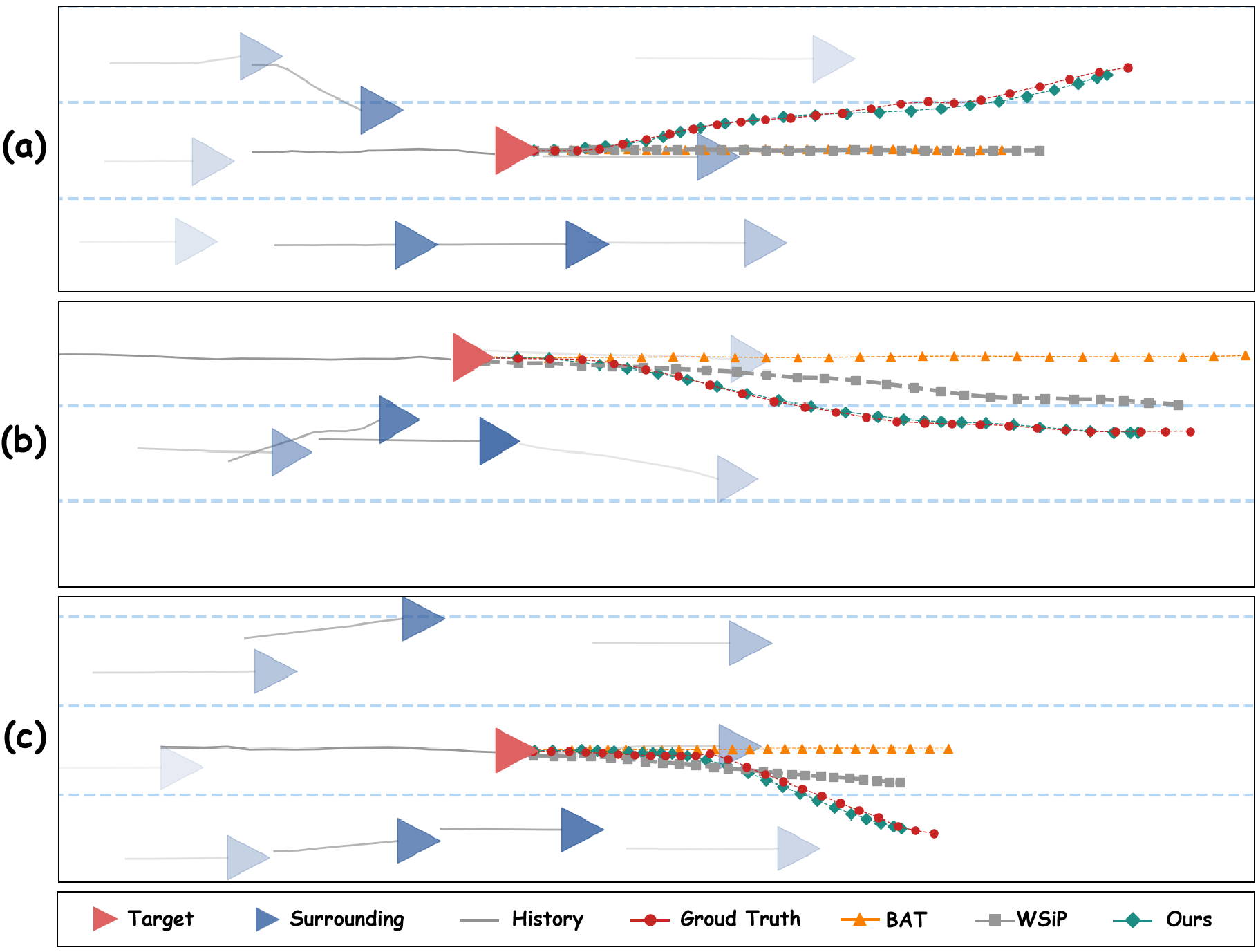} 
  \caption{{Visual insights from CITF and top baselines on the NGSIM dataset, illustrating short-term and long-term predictions for three complex driving scenarios: (a) merging, and (b-c) rightward lane change. A darker blue shade indicates an increased risk to the target vehicle, and vice versa.}}
  \label{case} 
\end{figure}

\subsection{Qualitative Results}
{
Figure \ref{case} compares the performance of CITF with the top baselines, BAT and WSiP, in complex highway scenarios. All three models perform well in short-term predictions, with predicted trajectories closely aligning with the ground truth. However, as shown in Figure \ref{case} (a), in the long-term prediction horizon, CITF (ours) successfully identifies the target vehicle's intent to merge into an adjacent lane and predicts the overtaking maneuver, producing an accurate forecast. In contrast, both the BAT and WSiP baselines incorrectly predict that the vehicle will continue driving straight. Moreover,
in Figures \ref{case} (b-c), facing the intricate dynamics and interactions typical of long-term predictions, such as sudden lane changes, surrounding traffic congestion, and potential collisions, CITF excels at capturing the subtle influences of driver behavior. It effectively models the traffic scene, recognizing the importance of surrounding vehicles and their impact on the target vehicle’s trajectory, thus maintaining high prediction accuracy. These visual results reveal that other models struggle to accurately discern vehicle intentions in complex long-term scenarios. This also further underscores the competitive performance of CITF in short-term predictions, as well as its outstanding capabilities in long-term prediction tasks.
}

\section{Conclusion}\label{Conclusion}
{On the journey to fully autonomous driving, long-term trajectory prediction remains a complex challenge. This study introduces an innovative approach rooted in cognitive insights, emphasizing the critical role of perceived safety in driver decision-making. Our Perceived Safety-Aware Module harmoniously merges Quantitative Safety Assessment and Driver Behavior Profiling, offering a detailed perspective on safety perceptions in driving. Rigorous evaluations on real-world driving datasets demonstrated the adaptability of our model, even under data constraints and missing data. In long-term prediction horizons, our proposed CITF significantly outperforms existing models, highlighting the promise of combining computational advantages with human cognitive processes to enhance both the safety and efficiency of autonomous driving. Despite significant progress in prediction accuracy and efficiency, the limitations of our model in short-term low-complexity scenarios warrant further investigation. In such scenes, minimal interactions between the target and surrounding vehicles make short-term dynamics largely governed by basic kinematic principles. A promising avenue for future research is to integrate physical models with deep learning techniques to better account for varying traffic scenes and improve the robustness of the model.}

\bibliographystyle{IEEEtran}

\bibliography{mybibfile}

%



%

\begin{IEEEbiography}
[{\includegraphics[width=1in,height=1.40in, clip,keepaspectratio]{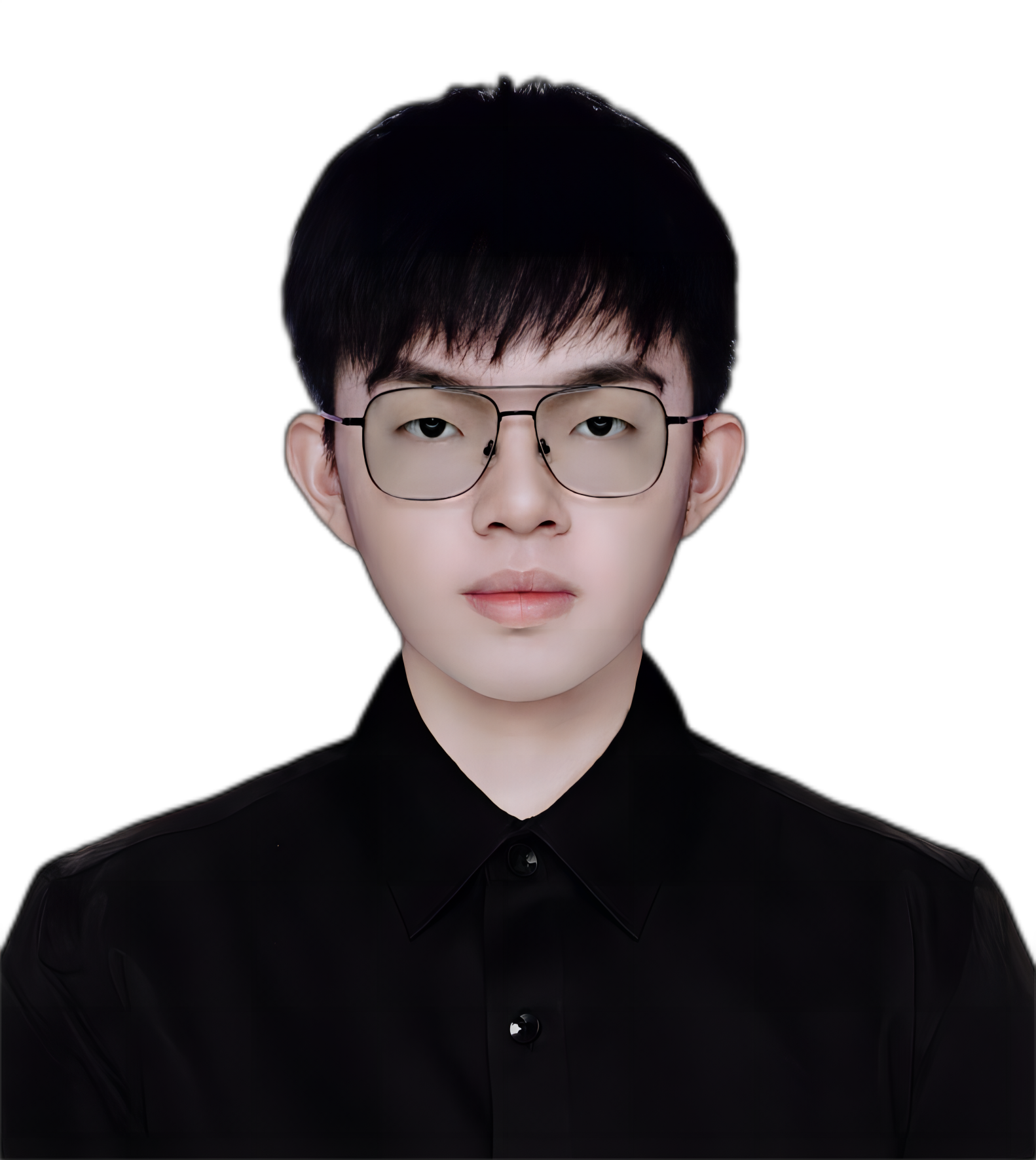}}]{Haicheng Liao} (Student Member, IEEE) received the B.S. degree in software engineering from the University of Electronic Science and Technology of China (UESTC) in 2022. He is currently pursuing the Ph.D. degree at the State Key Laboratory of Internet of Things for Smart City and the Department of Computer and Information Science, University of Macau. Over his academic career, he has published over 20 papers. His research interests include connected autonomous vehicles and the application of deep reinforcement learning to autonomous driving.
\end{IEEEbiography}

\begin{IEEEbiography}
[{\includegraphics[width=1in,height=1.40in, clip,keepaspectratio]{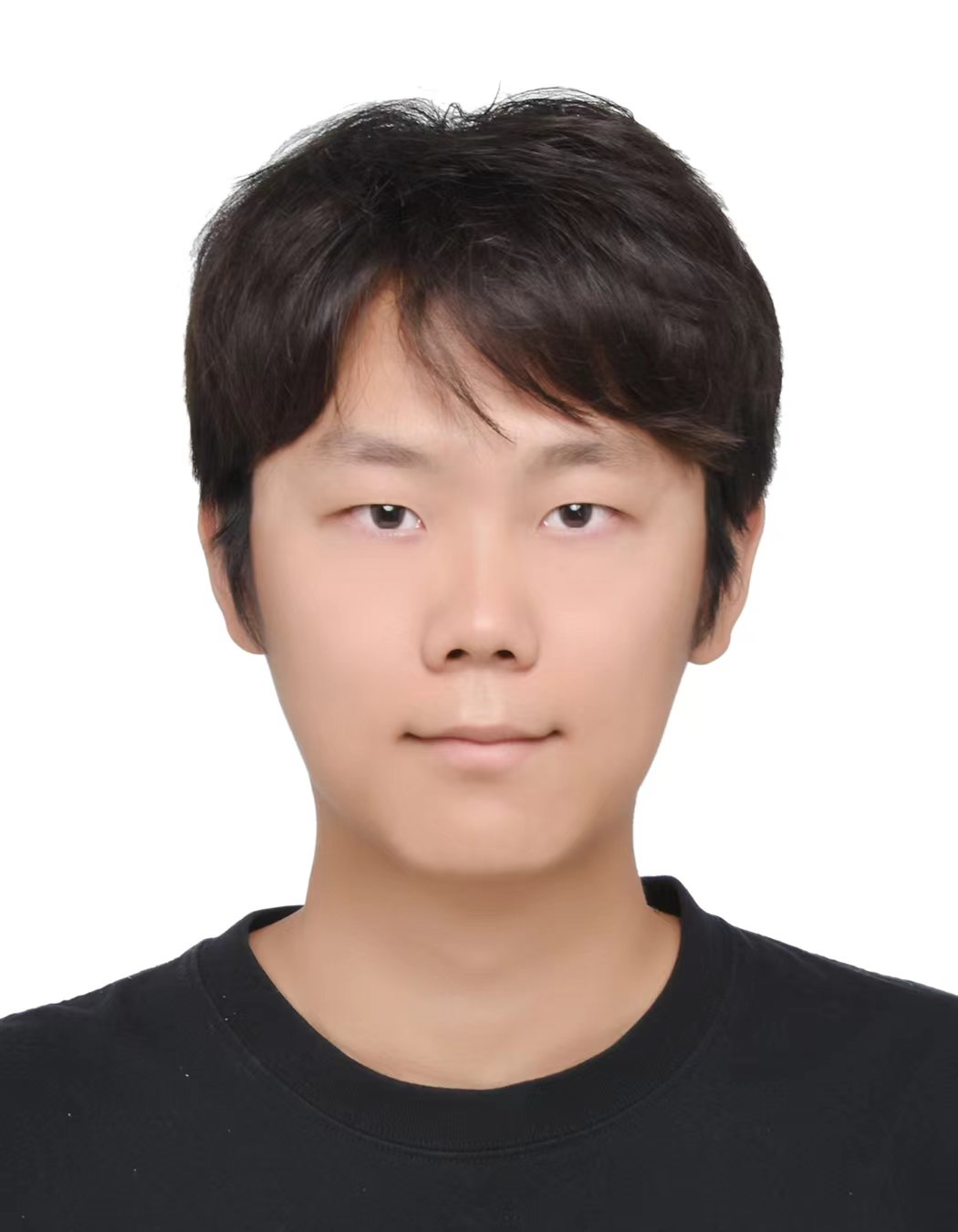}}]{Chengyue Wang} is currently pursuing a Ph.D. degree at the State Key Laboratory of Internet of Things for Smart City and the Department of Civil Engineering, University of Macau. He received his M.S. degree in civil engineering from the University of Illinois Urbana-Champaign (UIUC) in 2022. He received his B.E. degree in transportation engineering from Chang'an University in 2021. His research interests include connected autonomous vehicles and the application of deep reinforcement learning to autonomous driving.
\end{IEEEbiography}

\begin{IEEEbiography}
[{\includegraphics[width=1in,height=1.40in, clip,keepaspectratio]{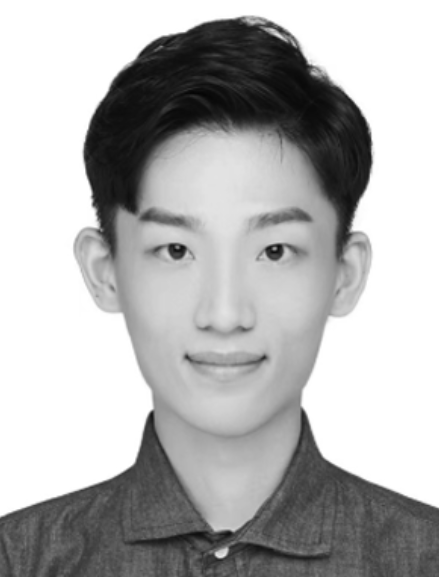}}]{Kaiqun Zhu} received the Ph.D. degree in control science and engineering from the University of Shanghai for Science and Technology, Shanghai, China, in 2022. From 2020 to 2022, he was a visiting Ph.D. student with the Department of Computer Science, Brunel University London, Uxbridge, U.K. He is currently a Postdoctoral Fellow with the University of Macau, Macau, China. His research interests include set-membership filtering, model predictive control, neural networks, privacy preserving, and their applications in autonomous vehicles.
\end{IEEEbiography}

\begin{IEEEbiography}
[{\includegraphics[width=1in,height=1.40in, clip,keepaspectratio]{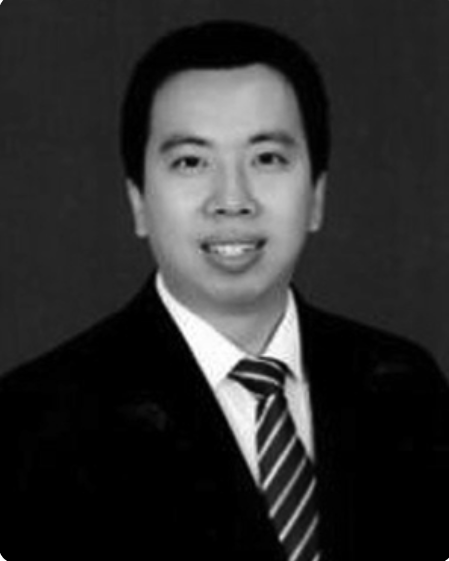}}]{Yilong Ren} (Member, IEEE) received the B.S. and Ph.D. degrees from Beihang University, Beijing, China, in 2010 and 2017, respectively. He is currently an Associate Professor with the Research Institute for Frontier Science, Beihang University. His research interests include vehicular communications, vehicular crowd sensing, and traffic Big Data.
\end{IEEEbiography}

\begin{IEEEbiography}
[{\includegraphics[width=1in,height=1.40in, clip,keepaspectratio]{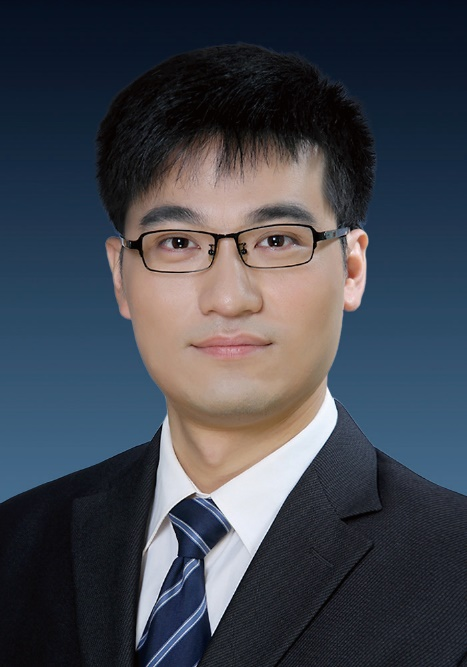}}]{Bolin Gao} received the B.S. and M.S. degrees in Vehicle Engineering from Jilin University, Changchun, China, in 2007 and 2009, respectively, and the Ph.D. degree in Vehicle Engineering from Tongji University, Shanghai, China, in 2013. He is now an associate research professor at the School of Vehicle and Mobility, Tsinghua University. His research interests include the theoretical research and engineering application of the dynamic design and control of intelligent and connected vehicles, especially about the collaborative perception and tracking method in cloud control system, intelligent predictive cruise control system on Commercial trucks with cloud control mode, as well as the test and evaluation of intelligent vehicle driving system.
\end{IEEEbiography}

\begin{IEEEbiography}
[{\includegraphics[width=1in,height=1.25in,clip,keepaspectratio]{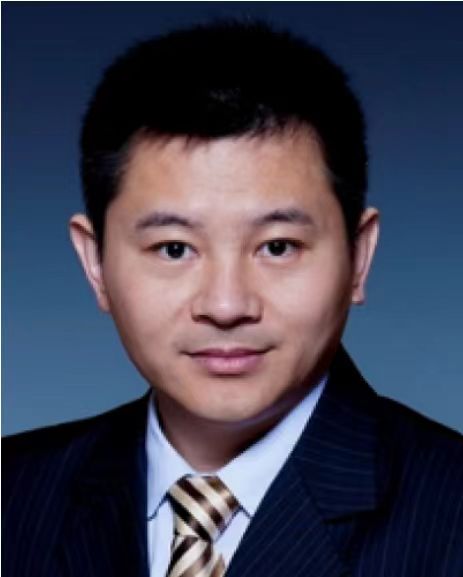}}]
{Shengbo Li} (Senior Member, IEEE) received his M.S. and Ph.D. degrees from Tsinghua University in 2006 and 2009, respectively. Before joining Tsinghua University, he had worked at Stanford University, University of Michigan, and UC Berkeley. His active research interests include intelligent vehicles and driver assistance, deep reinforcement learning, optimal control and estimation, etc. He is the author of over 130 peer-reviewed journal/conference papers and the co-inventor of over 30 patents. He is the recipient of the best (student) paper awards of IEEE ITSC, ICCAS, IEEE ICUS, CCCC, etc. His important awards include the National Award for Technological Invention of China (2013), the Excellent Young Scholar of NSF China (2016), the Young Professor of ChangJiang Scholar Program (2016), the National Award for Progress in Sci \& Tech of China (2018), Distinguished Young Scholar of Beijing NSF (2018), Youth Sci \& Tech Innovation Leader from MOST (2020), etc. He also serves as the Board of Governor of the IEEE ITS Society, Senior AE of IEEE OJ ITS, and AEs of IEEE ITSM, IEEE Trans ITS, Automotive Innovation, etc.
\end{IEEEbiography}

\begin{IEEEbiography}
[{\includegraphics[width=1in,height=1.25in,clip,keepaspectratio]{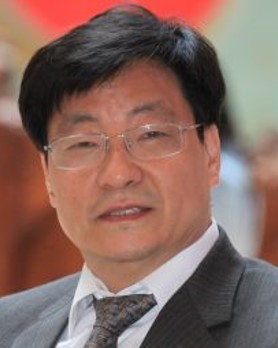}}]{Chengzhong Xu} (Fellow, IEEE) received the Ph.D. degree from The University of Hong Kong, in 1993. He is currently the chair professor of computer science and the dean with the Faculty of Science and Technology, University of Macau. Prior to this, he was with the faculty at Wayne State University, USA, and the Shenzhen Institutes of Advanced Technology, Chinese Academy of Sciences, China. He has published more than 400 papers and more than 100 patents. His research interests include cloud computing and data-driven intelligent applications. He was the Best Paper awardee or the Nominee of ICPP2005, HPCA2013, HPDC2013, Cluster2015, GPC2018, UIC2018, and AIMS2019. He also won the Best Paper award of SoCC2021. He was the Chair of the IEEE Technical Committee on Distributed Processing from 2015 to 2019.
\end{IEEEbiography}

\begin{IEEEbiography}
[{\includegraphics[width=1in,height=1.25in,clip,keepaspectratio]{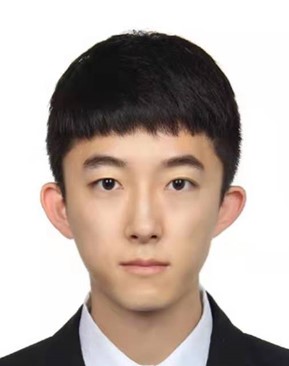}}] {Zhenning Li} (Member, IEEE) received his Ph.D. in Civil Engineering from the University of Hawaii at Manoa, Honolulu, Hawaii, USA, in 2019. Currently, he holds the position of Assistant Professor at the State Key Laboratory of Internet of Things for Smart City, as well as the Department of Computer and Information Science at the University of Macau, Macau.  His main areas of research focus on the intersection of connected autonomous vehicles and Big Data applications in urban transportation systems. He has been honored with several awards, including the Macau Science and Technology Award, Chinese Government Award for Outstanding Self-financed Students Abroad, TRB best young researcher award and the CICTP best paper award, amongst others.
\end{IEEEbiography}




\end{document}